# Raisonner avec des diagrammes :
# perspectives cognitives et computationnelles

## C. Recanati
L.I.P.N.
Université Paris 13
catherine.recanati@lipn.univ-paris13.fr

*Title:* « Reasoning with diagrams: cognitive and computational perspectives »

*Abstract:* Diagrammatic, analogical or iconic representations, are often contrasted with linguistic or logical representations, in which the shape of the symbols is arbitrary. The aim of this paper is to make a case for the usefulness of diagrams in inferential knowledge representation systems. Although commonly used, diagrams have for a long time suffered from the reputation of being only a heuristic tool or a mere support for intuition. The first part of this paper is an historical background paying tribute to the logicians, psychologists and computer scientists who put an end to this formal prejudice against diagrams. The second part is a discussion of their characteristics as opposed to those of linguistic forms. The last part is aimed at reviving the interest for heterogeneous representation systems including both linguistic and diagrammatic representations.

*Résumé* : On oppose souvent deux types de représentations, les représentations *diagrammatiques*, analogiques ou iconiques et les représentations *linguistiques* ou logiques, dans lesquelles la forme des symboles est arbitraire. Le but de cet article est de défendre l'intérêt des diagrammes dans les systèmes inférentiels de représentation de connaissance. Bien que largement utilisés, les diagrammes ont longtemps souffert de la réputation de n'être qu'un outil heuristique dans la recherche de solution, un simple support pour l'intuition. La première partie de cet article est un historique rendant hommage aux logiciens, psychologues et informaticiens qui ont mis un terme à ce préjugé formel contre les diagrammes. La deuxième partie discute les caractéristiques des diagrammes par opposition aux formes linguistiques dans le cadre des systèmes d'inférences. La dernière partie a pour but de raviver l'intérêt pour des systèmes de représentations hétérogènes comportant à la fois des représentations textuelles et diagrammatiques.

*Mots-clés* : diagrammes, raisonnement diagrammatique, représentations iconiques, représentations analogiques, système de représentation hétérogène, système inférentiel, modélisation cognitive.

# I. ARRIERE-PLAN HISTORIQUE

## I.1. Représentations diagrammatiques et représentations linguistiques

On oppose souvent deux types de représentations, les représentations *diagrammatiques*, analogiques ou iconiques et les représentations *linguistiques* ou logiques, dans lesquelles la forme des symboles est arbitraire – ce qui confère à ces systèmes, par opposition aux premiers, une plus grande universalité. Font partie du premier type : les figures en mathématique ou les schémas en physique, les icônes des interfaces homme/machine, les panneaux routiers, ou encore, certains symboles de notation musicale, etc. ; et il y a une grande variété de représentations iconiques différentes, allant des cartes géographiques aux schémas de composants électroniques, en passant par les cadrans horaires.

Mais la distinction entre les deux types de représentation sur la base de la forme des symboles n'est pas claire. Les systèmes iconiques utilisent des représentations dont la forme importe, mais également des symboles conventionnels et arbitraires, et toutes les relations spatiales présentes sur une figure n'ont pas nécessairement de signification (ni « la même » signification). Relativement au calcul, les systèmes logiques et linguistiques incorporent eux aussi une dimension géométrique, puisqu'ils manipulent des suites linéaires de symboles, où l'ordre et la place relative des termes sont essentiels[1]. Une partie des symboles utilisés, comme les flèches ou les parenthèses, y ont d'ailleurs un caractère iconique indéniable. Il y a donc différents degrés d'iconicité.

Comme le souligne Barbara Tversky (dans Tversky, 2003), bien avant le langage écrit, une multitude de marques picturales de toutes sortes ont existé : peintures rupestres, entailles sur des os, empreintes sur de l'argile, sur du bois. Ces éléments picturaux devaient être utilisés pour communiquer, garder trace d'événements dans le temps, enregistrer la propriété ou des transactions, situer des lieux, enregistrer des chansons ou des paroles, et dans beaucoup d'endroits du monde, l'utilisation de dessins pour communiquer s'est développée en de complets langages écrits.

Mais si les dessins utilisés avaient originairement une relation de ressemblance avec les objets qu'ils représentaient, tous ces langages ont dû trouver les moyens de représenter des choses abstraites, comme les noms propres, la quantification, la causalité, la négation, etc. Au fur et à mesure que les pictogrammes ont évolué en langages écrits, ils sont devenus plus schématiques et ont perdu peu à peu leur caractère iconique. Une telle métamorphose eut de formidables avantages : réduire le nombre de marques à apprendre, permettre de représenter des concepts abstraits (difficiles à schématiser par un dessin), et finalement permettre de représenter la parole elle-même. Les caractères sumériens par exemple, se transformèrent et utilisèrent les marques initiales pour schématiser des sons plutôt que les objets eux-mêmes.

Selon Tversky, avec l'avènement de l'alphabet, et plus tard de l'imprimerie, la transmission textuelle devint prépondérante dans la communication, et les figures plus décoratives que communicatives. Aujourd'hui, en partie à cause du développement des technologies pour créer, reproduire et transmettre des images, les dessins, figures et autres moyens de visualisations sont à nouveau en vedette. Mais l'augmentation des représentations picturales dans la communication est aussi due à la découverte, relativement récente, que ces représentations peuvent également véhiculer des sens abstraits au moyen de métaphores spatiales et visuelles. L'utilisation de l'espace permet en effet de capitaliser sur les considérables capacités humaines à faire des inférences spatiales.

---

[1] Dans les logiques intuitionnistes comme la logique linéaire, les règles sont bidimensionnelles et ont des propriétés géométriques (une preuve est identifiable à un réseau). Cependant, ces propriétés sont des « métapropriétés » indépendantes des propriétés du monde représenté, mais elles peuvent être utilisées pour guider ou fonder des démonstrations.

## I.2. Raisonner avec des diagrammes

Un des principaux soucis des fondateurs de la logique moderne était de justifier la validité des raisonnements par leur structure formelle. Pour Barwise et Etchemendy, ces pionniers, ayant tenu pour paradigmatiques les raisonnements valides en mathématiques, ouvrirent la voie aux développements que nous connaissons, et qui sont enseignés aujourd'hui en théorie des modèles, théorie de la preuve et théorie de la définissabilité. Ce modèle qui privilégiait les formes logiques *linguistiques* fut considéré a priori par certains de ses partisans (dont nombre d'informaticiens) comme intrinsèquement lié à cette forme particulière de représentation :

> The success of this model of inference led to an explosion of results and applications. But it also led most logicians – and those computer scientists most influenced by the logic tradition – to neglect forms of reasoning that did not fit well within this model. We are thinking, of course, of reasoning that uses devices like diagrams, graphs, charts, frames, nets, maps, and pictures.
>
> The attitude of the traditional logician to these forms of representation is evident in the following quotation, which expresses the standard view of the role of diagrams in geometrical proofs :
>
> « [The diagram] is only an heuristic to prompt certain trains of inference ; … it is dispensable as a proof-theoretic device ; indeed, … it has no proper place in the proof as such. For the proof is a syntactic object consisting only of sentences arranged in a finite and inspectable array ». (Neil Tennant, 1986, cité par B&E dans Barwise et Etchemendy, 1995).

Bien qu'étant largement utilisés dans les raisonnements, que ce soit pour résoudre des problèmes en physique, en mathématiques ou en logique, les diagrammes et les représentations visuelles en général, ont en effet souffert de la réputation de n'être qu'un outil heuristique dans la recherche de solutions, une sorte de simple support pour l'intuition. Or l'intuition est précisément ce dont la logique formelle dans la perspective de Frege entendait nous débarrasser.

Cependant, les diagrammes et autres objets de ce genre peuvent aussi être vus comme des objets syntaxiques pouvant se prêter à des raisonnements logiques dans une perspective formelle. Et si l'on adopte ce point de vue, le rejet des représentations diagrammatiques au profit des seules représentations linguistiques, peut apparaître comme un préjugé dont il convient de se défaire. Ce préjugé contre les diagrammes a été explicitement et largement exprimé par les mathématiciens, dont la méfiance provient selon Sun-joo Shin de deux sources principales : la première est la limitation intrinsèque des capacités représentationnelles des diagrammes, et la seconde, plus sérieuse, est la possibilité de leur mauvaise utilisation, qui peut induire, sans qu'on en ait conscience, des erreurs dans les raisonnements (Shin, 1994). Cependant, historiquement, personne n'a jamais donné de justification théorique à ce préjugé, qui n'a pas non plus été mis en cause par les partisans des diagrammes qui se sont contentés d'argumenter en leur faveur en montrant leurs avantages dans des applications.

Dans cet article, nous commencerons par mentionner les travaux théoriques qui ont contribué à désamorcer ce préjugé des logiciens contre les diagrammes, puis nous tâcherons de faire le point sur leurs caractéristiques, en soulignant principalement ce qui les distingue des textes. Nous conclurons en mentionnant leurs avantages dans les systèmes inférentiels hybrides car nous avons aujourd'hui l'intime conviction que *seuls des systèmes de représentation hybrides ou hétérogènes (notés SRH),* c'est-à-dire articulant des sous-systèmes de représentation de différentes natures, logiques et analogiques, peuvent permettre de construire des modèles de raisonnement *computationnellement* et *cognitivement* plausibles.

Quoique les diagrammes aient pu ainsi être réhabilités en partie par et grâce aux travaux de logiciens, c'est aux informaticiens que l'on doit la plupart des contributions qui ont montré leur intérêt. En Intelligence Artificielle, de brillants auteurs comme McCarthy et Hayes, avaient soutenu dans les années 50-60, malgré la perplexité de nombre de leurs contemporains, que des systèmes fondés uniquement sur des représentations logiques devaient pouvoir manifester des capacités proches de l'intelligence humaine. Ce fut alors la grande époque de Lisp, Prolog, et du développement des systèmes experts. Au cours des années 70, ce dogme « logiciste » fut mis en doute, car il était interprété de manière restrictive, comme lié aux formes des représentations utilisées, et un débat théorique assez vif s'ensuivit. De nombreux informaticiens prônaient en effet à l'inverse l'utilisation de représentations spécifiques, adaptées à la fois au problème et au domaine. Des débats parallèles eurent lieu en psychologie de la vision entre les partisans de l'imagerie mentale et leurs opposants, et, en psychologie du raisonnement, entre partisans de l'existence d'une logique mentale et partisans de modèles mentaux. En Intelligence Artificielle, le débat s'augmenta de questions sur les rapports entre le pouvoir expressif d'un langage ou d'un système de représentation et ses possibilités calculatoires (voir, l'étude comparative des différents formalismes de représentations de connaissances dans Brachman et Levesque, 1985).

Le philosophe Aaron Sloman fut l'un des premiers en 1971 à manifester des doutes sur l'affirmation selon laquelle la logique devait suffire au projet de l'I.A. Pour sa vision historique de la querelle, voir Sloman 1995. Selon lui, des débats identiques resurgissent périodiquement, et les arguments d'alors sont encore agités aujourd'hui, reproduisant le même type d'erreurs et de confusions.

Nous pensons avec lui que la question de fond concernant l'intelligence reste une question d'architecture et d'articulation entre systèmes de représentation, et qu'elle doit passer pour être appréhendée correctement, par une étude préliminaire des caractéristiques des différentes formes de systèmes de représentation. Ainsi les travaux sur les raisonnements diagrammatiques permettent de mettre en évidence certaines de ces caractéristiques, mais il reste évidemment encore beaucoup à faire pour comprendre comment l'information peut être mémorisée et manipulée dans des systèmes intelligents, qu'ils soient naturels ou artificiels.

## II. FONDEMENTS THEORIQUES

La contribution de Jon Barwise et John Etchemendy est sans aucun doute la plus importante et la plus prometteuse des contributions dans ce domaine. L'intérêt et l'originalité de leur approche est qu'elle est fondamentalement mathématique et logique, et qu'elle s'oppose résolument à une conception restreinte de la logique, qui stipule que cette dernière devrait se cantonner à ne manipuler que des formes linguistiques.

Selon B&E, la vision générale ayant émergé de la logique traditionnelle – dans laquelle la grande majorité des raisonnements considérés comme valides, sinon tous, serait effectuée dans le langage, et qui prétend que cette sorte de raisonnement serait aujourd'hui bien compris – est fausse à double titre : elle est évidemment fausse en ce qui concerne la fréquence de l'utilisation de formes non linguistiques dans les raisonnements ; et ensuite, les théories logiques actuelles de l'inférence ne rendent de toutes manières pas compte des raisonnements basés sur des formes linguistiques.

> As research in semantics over the past twenty years has shown, human languages are infinitely richer and more subtle than the formal languages for which we have anything like a complete account of inference. […] we think it is important for logicians to broaden their outlook beyond linguistically presented information. As the computer gives us ever richer tools for representing information, we must begin to study the logical aspects of reasoning that uses nonlinguistic forms of representation. In this way we can hope to do something analogous to what Frege and his followers did for reasoning based on linguistic information. […] Our hope is that the tools we have begun to develop will allow something similar to be done with information presented in more than one mode. (Barwise et Etchemendy, 1990 b).

Cette conviction, qu'une théorie générale des inférences valides peut-être développée indépendamment des modes de représentations de l'information, est basée sur leur vision *informationnelle* de la logique : une règle d'inférence est valide si elle garantit que l'information représentée dans la conclusion est implicite dans celle représentée dans les prémisses.

La théorie générale des situations que J. Barwise cherchait à élaborer était un cadre mathématique pour rendre compte de l'aspect profondément réflexif de la modélisation des situations du monde. Les situations de Barwise (par exemple une phrase dans une conversation, un programme d'ordinateur, une main au poker, ou un avion du trafic national aérien), sont en effet des composants de systèmes complexes qui sont eux-mêmes des systèmes complexes[2]. Dans de tels cadres, il s'attachait à définir comment l'information pouvait être transmise logiquement, d'un système supportant une logique « locale » à un autre (Barwise et Etchemendy, 1990 a).

Les situations devant permettre de modéliser les systèmes complexes, on ne s'étonnera pas que Barwise ait eu le souci de légitimer l'utilisation de représentations visuelles dans les raisonnements. Il est en effet très difficile de définir de tels systèmes sans faire un usage important de diagrammes ou de schémas. C'est sous sa direction que Sun-joo Shin a fait sa thèse à Stanford en 1991. Cette thèse est remarquable car on y trouve enfin la démonstration que des systèmes utilisant des représentations diagrammatiques, en l'occurrence ici des diagrammes inspirés de ceux de Venn et de Peirce, peuvent devenir l'objet d'une démonstration rigoureuse, établissant qu'ils constituent des systèmes déductifs corrects et complets. Pour faire cette démonstration, Shin munit ces systèmes d'une syntaxe (établissant la notion de diagrammes bien formés), d'une sémantique (ici le système entend représenter des ensembles au moyen de régions et permettre la résolution de syllogismes constitués d'assertions sur des ensembles), et de règles de transformations diagrammatiques ou d'axiomes (permettant de définir comment un diagramme dérive d'un autre, est équivalent à un autre, et comment on peut obtenir un diagramme à partir de plusieurs autres, etc.). La seule étude de ce type l'ayant précédée est due à John Sowa qui donna en 1984 un traitement formel des graphes existentiels de Peirce incluant leur syntaxe, sémantique et théorie de la preuve[3] (Sowa, 1984).

La première partie de la thèse de S. Shin présente en détail les différents apports des diagrammes d'Euler, de Venn et de Peirce, pour la résolution de syllogismes. L'utilisation de cercles pour illustrer des relations entre des classes est généralement attribuée à Euler[4] (Euler, 1768). Mais le fait que ses diagrammes ne permettaient pas d'exprimer toutes les relations entre deux classes simultanément en limitait considérablement le pouvoir expressif et déductif. Au dix-neuvième siècle, sa méthode fut modifiée et améliorée par Venn (Venn, 1894 ou 1971) qui

---

[2] Un système complexe est défini selon Barwise par le fait qu'il implique des composants distribués dans le temps et l'espace, de multiples agents, de multiples formes de représentations et différentes sortes de régularités.

[3] Son formalisme est équivalent à celui de la logique du premier ordre, mais il est donné ici en termes diagrammatiques.

[4] Un des rapporteurs de cet article m'a signalé que contrairement à ce qu'écrivent la plupart des auteurs sur les diagrammes, Euler n'est pas le premier à avoir proposé les diagrammes qui portent son nom. Il fut précédé par d'autres, et en particulier par Leibniz un siècle plus tôt. Ce dernier a proposé (outre un système de lignes) exactement les mêmes quatre diagrammes qu'Euler. On trouve le texte en latin dans Couturat, 1903, mais cet opuscule n'a pu être daté avec précision.

réussit à faire figurer sur un même diagramme toutes les relations possibles entre deux classes. Mais Venn ne se posa pas la question des existentielles. Il se contentait de représenter les classes vides. Ainsi, l'assertion « Tous les A sont des B » véhiculait pour lui l'information que la classe des A qui ne sont pas des B est vide, et il hachurait sur son diagramme la région correspondante :

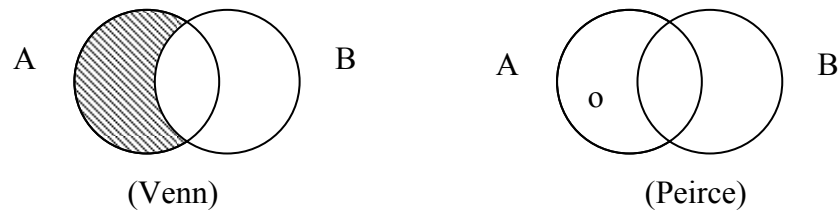

(Venn)    (Peirce)

Figure 1. Tous les A sont des B

L'apport de Peirce vingt ans plus tard fut d'ajouter la possibilité d'exprimer des assertions existentielles (indiquées par une croix) et des disjonctions (indiquées par un trait permettant de lier des zones) – les zones vides se trouvant annotées par un simple rond, marquant la négation de l'existence (Peirce, 1933).

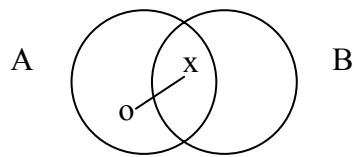

Figure 2. Tous les A sont des B ou Il existe un A qui est un B (Peirce)

C'est également à Peirce que l'on doit l'idée de règles de transformations sur les diagrammes. Mais, la question de savoir si ces règles sont valides ou non nécessitait encore un pas supplémentaire, et c'est ce travail qu'a réalisé S. Shin. En utilisant la distinction moderne entre syntaxe et sémantique, elle donne la formalisation complète de deux systèmes de déduction diagrammatiques, permettant de tester la validité de formes restreintes de syllogismes. Le premier système permet de tester des syllogismes catégoriques[5]. Le deuxième permet d'étendre le premier à la résolution de syllogismes dont les prémisses et les conclusions sont des assertions catégoriques étendues, i.e. pouvant être constituées d'assertions catégoriques reliées par des connecteurs logiques. Elle montre ainsi que les représentations diagrammatiques du premier système peuvent encore, moyennant certaines extensions, être rendues équivalentes à un langage monadique du premier ordre. Mais, si l'entreprise est intéressante d'un point de vue didactique, le second système fait usage de diagrammes multiples et il n'a pas la simplicité du premier. La recherche de systèmes hétérogènes, qui manipulent à la fois des représentations linguistiques et des représentations diagrammatiques, pouvant ainsi tirer parti des avantages des deux types de représentations, est en réalité bien préférable[6].

C'est dans cet état d'esprit que Barwise et Etchemendy ont argumenté contre l'idée d'un système universel de représentation diagrammatique, mais également, et de manière plus intéressante, contre l'idée naturellement répandue, qu'un système utilisant de multiples formes de représentations aurait besoin d'un langage intermédiaire pour relier ces différentes représentations (cf. Barwise et Etchemendy, 1995). Les anciens partisans de la logique traditionnelle sont sans doute prêts à se laisser convaincre de la possibilité d'étendre des propriétés déductives à des sous-systèmes manipulant des représentations non linguistiques, et cela dans un but d'efficacité, mais ils pourraient cependant encore rester convaincus que l'articulation finale de ces sous-systèmes ne peut être effectuée correctement que dans un langage logique traditionnel commun.

---

[5] Un syllogisme catégorique consiste en deux prémisses et une conclusion – lesquelles sont constituées de phrases catégoriques simples, qui par définition ne contiennent pas d'implication ou de disjonction.

[6] Et cela, pour diverses raisons : en premier lieu, pour des raisons d'efficacité mais aussi pour la clarté et la modularité des modèles ainsi proposés (plus facilement compréhensibles, communicables et modifiables) ; et d'autre part pour des raisons de plausibilité, dans des cas particuliers de modélisation cognitive.

Mais ce qui fonde théoriquement l'articulation des différents sous-systèmes de représentation d'un système complexe est pour B&E simplement *le fait qu'ils dénotent les mêmes objets du monde*. Ainsi, deux sous-systèmes peuvent dénoter des propriétés différentes, et ce qui est exprimable dans un sous-système peut ne pas l'être dans l'autre. Certaines informations peuvent néanmoins être transférées d'un sous-système à l'autre, donnant à l'ensemble des capacités inférentielles supérieures. Le développement d'un langage global permettant de transférer toute information d'un sous-système vers un autre n'est pas absolument nécessaire. Dans Barwise et Etchemendy 1995, B&E fournissent en effet une justification théorique des deux algorithmes implantés dans *Hyperproof*[7] pour transférer des informations entre diagrammes et formules - justification mathématique qui ne fait pas usage d'un langage intermédiaire.

Mais pour pouvoir tirer le meilleur parti des systèmes de représentation hétérogènes, il est important d'abord de bien comprendre les propriétés intrinsèques des différents types de systèmes de représentation. Dans la section suivante, nous allons donc tenter de faire le point sur les propriétés spécifiques des systèmes diagrammatiques, par opposition aux systèmes logiques linguistiques.

## III. QU'EST-CE QUI DISTINGUE LES SYSTEMES DIAGRAMMATIQUES ?

### III.1. Clôture sous contraintes et homomorphisme syntaxique

Dans Barwise et Etchemendy (1990 b), B&E relèvent cinq caractéristiques relativement fréquentes des raisonnements comportant des figures. Les systèmes diagrammatiques manifestent (1) une clôture sous contraintes, (2) la conjonction plutôt que la disjonction d'information, (3) des représentations homéomorphes, et ces représentations supportent (4) des arguments de symétries, et (5) des inférences perceptuelles. Nous pensons que les trois premières caractéristiques sont l'expression d'une même propriété. La quatrième nous paraît relativement contingente, et nous discuterons la cinquième dans une autre section.

Le caractère *clos sous contraintes* des représentations apparaît selon B&E lorsque les contraintes sur un diagramme coïncident avec les contraintes sur une situation. Le diagramme peut alors générer beaucoup d'informations sans que l'utilisateur ait besoin de les inférer. L'utilisateur peut en effet *lire directement des faits* (conséquences des faits initiaux et des contraintes de la situation) sur le diagramme, par *simple inspection*. C'est comme si les conséquences des faits initiaux étaient nécessairement incluses dans la représentation sous l'effet de contraintes structurelles. Cette situation est en contraste brutal avec celle des inférences linguistiques, où la plus triviale conséquence doit être inférée explicitement.

---

[7] *Hyperproof* est un logiciel interactif, destiné à l'enseignement de la logique aux étudiants ; il dépeint un monde de formes géométriques disposées sur les cases d'un damier dans deux formats de représentations différents : un langage logique du premier ordre, et une représentation picturale. L'utilisateur du logiciel peut, étant donnée la description d'un monde fournie par la conjonction d'informations données dans les deux modes, vérifier interactivement si une nouvelle information donnée est valide ou contradictoire, en faisant une preuve « hybride ».

Perry et Macken paraphrasent cette propriété par :

> Constraints on the facts in a representation R that represent facts about a relation Q are such that IF Q-facts $f_1 \ldots f_n$ guarantee f-facts $f_{n+1}$, THEN R will explicitly represent $f_{n+1}$. (Perry et Macken, 1996).

Ils critiquent alors le caractère non suffisant de cette condition en argumentant qu'une représentation linguistique dotée d'un mécanisme interne qui ajouterait automatiquement à la représentation de tout nouveau fait, toutes les représentations des conséquences logiques de l'adjonction de ce nouveau fait, serait alors une représentation fermée sous contraintes, mais pas diagrammatique pour autant.

Cette critique de P&M va nous permettre de mieux comprendre la caractéristique isolée par B&E. Pour être acceptable, la reformulation que P&M ont proposée imposerait en effet de préciser ce que le terme d'« explicitement » signifie. Un fait important que souligne la caractéristique de B&E – et qui est selon nous une caractérisation primordiale des raisonnements diagrammatiques par opposition aux raisonnements logiques linguistiques – n'est en effet pas correctement rendu par cette formulation, en tout cas, telle que P&M l'entendent.

Dans un système de représentations clos sous contraintes, on se contente de « représenter » les faits en les ajoutant un à un à un sur un diagramme, et on peut ensuite observer directement sur le diagramme obtenu les conséquences logiques de ces faits, *sans avoir à exécuter de calcul*. Car c'est dans des contraintes quasi « physiques » (sur les représentations elles-mêmes) que sont cachées les contraintes logiques abstraites de la situation représentée.

Dans un système de représentation linguistique, on énumère les faits, puis on fait un calcul pour en dériver une conséquence.

Ainsi pour reprendre le contre-exemple de P&M, la situation de trois figures géométriques alignées (un triangle T, un losange L et un carré C), peut être représentée diagrammatiquement par l'alignement de trois lettres, T, L et C, symbolisant les trois figures et leurs positions relatives :

$$T \ L \ C$$

Figure 3. Diagramme des 3 figures alignées

A l'inverse, une description linguistique basée sur des phrases de la forme « X est à gauche de Y » permettrait de décrire la même situation avec les deux phrases

> T est à gauche de L
> L est à gauche de C

Figure 4. Représentation linguistique

La reformulation de la clôture sous contraintes de P&M prétend qu'il suffirait de rajouter systématiquement à toute adjonction d'un nouveau fait ses conséquences logiques, pour obtenir une représentation fermée sous contraintes. De sorte que, si l'on ajoutait à la représentation de «T est à gauche de L» la représentation du fait que «L est à gauche de C», on ajouterait (grâce à un mécanisme inférentiel interne) non seulement la phrase «L est à gauche de C», mais également la représentation des conséquences logiques de son adjonction, ici «T est à gauche de C». La situation décrite par la figure serait donc dans ce système représentée linguistiquement par les trois phrases « T est à gauche de L », « L est à gauche de C » et « T est à gauche de C », où toutes les conséquences des faits sont explicites. Mais, argumentent P&M, cela ne rendrait pas cette représentation diagrammatique pour autant[8], et la propriété de clôture sous contraintes n'est donc pas suffisante.

P&M ont raison d'isoler la propriété logique ainsi reformulée, car elle est effectivement distincte de la propriété décrite par B&E. Mais si l'on considère que ce qui est *explicite* dans une représentation est *ce qui ne nécessite pas de calcul* (le calcul étant ici un passage de l'implicite à l'explicite) le contre-exemple de P&M n'est plus recevable. On peut en effet souligner que dans ce contre-exemple, la propriété de transitivité de la relation « à gauche de » aura besoin d'être explicitement fournie dans le nouveau système, et qu'elle serait utilisée par le mécanisme inférentiel interne pour produire la représentation du nouveau fait. Le coût de l'ajout de la nouvelle représentation serait donc en réalité augmenté du coût d'un calcul, celui précisément de ses conséquences, et la représentation considérée ne serait pas close sous contraintes.

L'exemple a cependant le mérite de souligner cette profonde dualité entre raisonnement diagrammatique et raisonnement linguistique par rapport au calcul. Un raisonnement logique peut en effet être conceptualisé comme la représentation de faits explicites permettant d'établir une conséquence implicite. Il y a donc deux phases : une phase établissant la représentation de faits explicites, et une seconde phase, rendant explicite un autre fait à partir de cette représentation. Dans les raisonnements linguistiques, tout le poids du calcul est porté par la deuxième phase, alors que dans les raisonnements diagrammatiques, cette phase n'existe pas. Cette dualité est l'expression du dilemme entre calcul et représentation. Ce dilemme est bien connu en informatique, où il prend souvent la forme de la distinction entre programme et données, et en représentation des connaissances, entre connaissances déclaratives et connaissances procédurales.

Ainsi, les raisonnements logiques nécessitent (1) des représentations linguistiques explicites de faits (souvent indépendantes et généralement selon une dimension simple de spécificité) et (2) des représentations linguistiques explicites des propriétés abstraites ou des relations entre les différentes dimensions. Pour établir la validité d'une conséquence logique non explicite au départ, ils nécessitent alors le déroulement d'un calcul articulant les deux sources de représentations.

A l'inverse, les raisonnements diagrammatiques dans un système clos sous contraintes ont ceci de particulier que la structure même des représentations contraint l'information à être explicitée. Pour cela : (1) la représentation des faits (même selon une dimension simple de spécificité) n'est pas fournie de manière isolée, mais à l'intérieur d'une structure globale. Elle s'effectue donc toujours de manière contextuelle, contrairement au cas logique ; et (2) Le système *ne nécessite pas* la représentation explicite de propriétés plus abstraites ou de contraintes entre propriétés. Celles-ci sont implicitement prises en compte par les propriétés de la structure englobante. Pour établir la validité d'une conséquence logique non donnée initialement, ils n'imposent pas, de facto, de calcul articulant deux sources de représentations, puisqu'il n'y en a qu'une ! Dans ce type de système, on ne vérifie pas la validité d'une conséquence logique : on inspecte seulement la représentation pour vérifier que la conséquence recherchée s'y trouve représentée.

Pour conclure cette discussion sur la clôture sous contraintes de B&E, on pourrait la formuler de la façon suivante (en utilisant en réalité le troisième critère de B&E) :

---

[8] Comme me l'a fait remarquer D. Kayser, on n'a pas réellement besoin ici d'une situation à trois figures, on peut d'emblée dire que la dérivation 'L est à droite de T' nécessite un « calcul » dans le cas linguistique, alors qu'elle est « perceptuelle » dans le cas diagrammatique.

Un système de représentation diagrammatique est *clos sous contraintes* ssi il existe un homomorphisme entre les contraintes sur les représentations et celles de la situation représentée, tel que toutes les conséquences logiques de la situation représentée soient explicites dans la représentation.

De cette propriété dérive une propriété syntaxique très forte. Dans les systèmes de représentation diagrammatiques clos sous contraintes, la satisfaction des contraintes structurelles de la situation est nécessaire à toute représentation bien formée. *Les contradictions n'y sont donc pas représentables, et tout diagramme bien formé correspond de façon nécessaire à une situation logiquement possible*. Cette propriété est véritablement une caractéristique propre des systèmes diagrammatiques (prototypiques) par opposition aux systèmes linguistiques, dans lesquels la contradiction n'est pas marquée syntaxiquement.

Elle permet en outre d'expliquer la seconde caractéristique de B&E, car le traitement aisé des conjonctions et malaisé des disjonctions dans les systèmes diagrammatiques en dérive souvent.

En effet, dans un système clos sous contraintes, il suffit pour représenter la conjonction d'un nouveau fait et d'une ensemble de faits, d'ajouter la représentation du nouveau fait au sein de la représentation globale correspondant à cet ensemble de faits. En revanche, le traitement diagrammatique de cas disjonctifs amène nécessairement à considérer des diagrammes disjoints. C'est ce qui se produit dans l'exemple des cinq chaises donné par B&E :

> You are to seat four people, A, B, C and D in a row of five chairs. A and C are to flank the empty chair. C must be closer to the center than D, who is to sit next to B. From this information, show that the empty chair is not in the middle or on either end. Can you tell who is to be seated on the two ends ? (Barwise et Etchemendy, 1990 b).

La représentation diagrammatique d'une configuration proposée par B&E consiste à représenter les cinq chaises par cinq traits horizontaux consécutifs, et à disposer les quatre personnages A, B, C et D au-dessus de ces traits pour indiquer « qui » est assis « où ». La chaise vide quant à elle est surmontée d'une croix. Le raisonnement s'appuie alors sur l'inspection des trois diagrammes permettant de rendre compte de la première hypothèse (Figure 5), et il conduit aux deux possibilités finales représentées Figure 6, après élimination des cas 2 et 3.

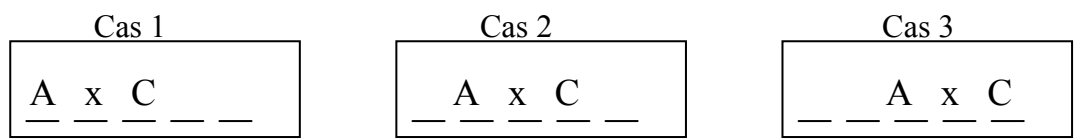

Figure 5. Les trois premiers cas

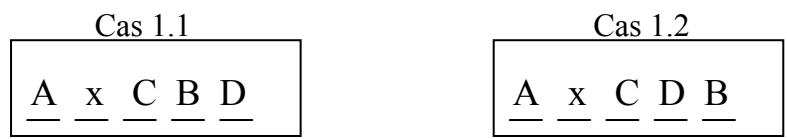

Figure 6. Les deux possibilités finales

La deuxième hypothèse, stipulant que C doit se trouver plus près du centre que D élimine en effet le cas 3, et la troisième hypothèse, qui stipule que D doit se trouver assis à côté de B, élimine le cas 2. On notera au passage que ce sont les contraintes créées par le positionnement de A et C dans les cas 2 et 3 qui conduisent à rendre impossible la représentation des autres hypothèses dans ce contexte[9].

La difficulté à traiter la disjonction est ici une conséquence directe de la clôture sous contraintes[10]. Deux cas antinomiques ne peuvent en effet pas être représentés sur un même diagramme puisqu'un diagramme bien formé ne peut représenter de situation logiquement contradictoire. La propriété de clôture sous contraintes apporte donc aussi avec elle une importante limitation aux diagrammes : si on doit raisonner sur des cas exclusifs dans un système fermé sous contraintes, on perd alors l'un des principaux avantages de ce type de système, qui est d'avoir toutes les conséquences explicites au sein *du même diagramme*. Si plusieurs cas doivent être envisagés, il faudra considérer plusieurs diagrammes, et donc, pour vérifier qu'une conséquence est valide, parcourir l'ensemble des diagrammes et vérifier la validité dans chaque cas.

Néanmoins, contrairement à ce que beaucoup d'auteurs ont soutenu, il est possible de représenter diagrammatiquement des disjonctions sur un diagramme, en introduisant des symboles du second ordre. Peirce le premier a proposé de représenter des existentielles sous la forme de deux symboles reliés par un trait, pour indiquer l'existence d'un élément dans l'une ou l'autre de deux régions sur un diagramme. Mais l'introduction de symbole du second ordre nécessite l'introduction de règles « logiques » de transformations sur les diagrammes. Il faut alors introduire une notion de diagrammes équivalents, une notion d'implication entre diagrammes, etc. On fait alors de véritables raisonnements sur les diagrammes, en y réalisant des opérations graphiques légitimes, mais on perd l'avantage des systèmes clos sous contraintes, puisqu'il faut à nouveau « calculer » pour rendre un résultat explicite. Mais comme l'a montré S. Shin, ces raisonnements peuvent cependant être valides et logiquement fondés, car les notions de correction et de complétude s'appliquent aussi bien aux systèmes diagrammatiques qu'aux systèmes linguistiques.

Il reste donc vrai que les représentations diagrammatiques sont rarement utilisées pour figurer des disjonctions, et de manière générale, qu'elles sont optimales dans le traitement des conjonctions car la propriété de clôture sous contraintes peut alors s'y manifester.

Pour conclure sur le troisième critère de B&E, il faut préciser, comme le remarquent P&M, que l'existence d'un homomorphisme n'est pas en soi une caractéristique des systèmes de représentation diagrammatiques : on peut toujours définir un homomorphisme entre une description textuelle décrivant une situation d'occupation de chaises (Figure 7), et une situation du monde. Il suffit d'utiliser l'homomorphisme induit naturellement entre le langage de description des faits *représentants* et le langage de description des faits *représentés*, comme le

---

[9] On utilise en outre [...] es positions de A et C. Ce type d'argume[nt ...] diagrammes, et souvent, comme dans ce[...] tte fréquence est bien entendu liée au ca[...] s relativement à l'« essence » des raiso[nnements ...] ristique mineure qui n'est ni nécessaire n[i ...] manière dont les symboles sont utilisés p[our ...] ourcir la preuve, ils ne sont pas caractéri[stiques ...] instance, sur un diagramme proprement [...] alogues, comme des permutations sur des termes, pour raccourcir les preuves dans des raisonnements logiques ou mathématiques, et nous ne pensons pas que ce point mérite réellement d'être retenu.

$a \neq b$, $a \neq c$, etc.
a est une chaise, b est une chaise, etc.
a est à gauche de b, b est à gauche de c, etc.
A est une personne, B est une personne, etc.
A est assis sur a, B est assis sur c, etc.
b est vide.

[10] Cela donne une certaine plausibilité à la thèse que les représentations cognitives sont en grande partie diagrammatiques, car les humains raisonnent mieux sur des conjonctions que sur des disjonctions (Bruner et alii, 1956).

montre la mise en correspondance des deux colonnes du tableau de la Figure 8 (Perry et Macken, 1996).

Figure 7. Une représentation textuelle d'une situation de chaises

Le troisième critère de B&E présuppose donc une interprétation particulière des termes de « représentations homéomorphes ». Pour B&E, il s'agit en réalité de représentations qui se correspondent dans un homomorphisme qui conserve fortement des relations structurelles, (et c'est d'une telle propriété de conservation forte que dérive la clôture sous contraintes).

Dans « Heterogenous Logic » (Barwise et Etchemendy, 1995) ils énumèrent sept caractéristiques des systèmes de représentation possédant des représentations « homéomorphes » à la situation représentée. Mais il ne s'agit pas pour eux de caractéristiques nécessaires ou suffisantes, et un système de représentation peut présenter divers degrés d'homomorphie, qu'il soit diagrammatique ou linguistique.

Parmi ces traits, ils relèvent que dans les systèmes de représentation diagrammatiques, les caractéristiques suivantes sont fréquentes :

1. Les objets sont représentés par des exemplaires d'icônes. Il est fréquent que chaque objet soit représenté par un exemplaire unique et que des exemplaires distincts représentent des objets distincts.

| **Situation de chaises** | **Description textuelle** |
|---|---|
| Pas les mêmes chaises, a, b | Les symboles «a» et «b» sont flanqués du symbole «≠» |
| etc. | etc. |
| a est une chaise | Le symbole «est une chaise» précède le symbole «a» |
| etc. | etc. |
| A est une personne | Le symbole «A» précède le symbole «est une personne» |
| etc. | etc. |
| a est à gauche de b | Les symboles «a» et «b» sont flanqués de «est à gauche de» |
| etc. | etc. |
| A est assis sur a | Les symboles «A» et «a» sont flanqués de «est assis sur» |
| etc. | etc. |
| b n'est pas occupée | Le symbole «est vide» suit le symbole «b» |

Figure 8. Homomorphisme entre description textuelle et situation de chaises

2. Il existe une correspondance entre les types d'icônes et les propriétés des objets qui

préserve la structure. On peut s'attendre à ce que parfois :

(a) Si un type d'icône est un sous-type d'un autre, alors il y a aussi une relation de sous-propriété entre les propriétés ainsi représentées.

(b) Si deux types d'icônes sont incompatibles, alors les propriétés qu'ils représentent devraient être incompatibles.

Mais le contraire de (a) et (b) peut aussi bien se produire.

3. Les propriétés d'ordre supérieur des relations entre objets sont représentées par les mêmes propriétés relationnelles entre les exemplaires d'icônes.
4. Chaque possibilité, concernant les objets représentés, leurs propriétés et leurs relations, est représentable, i.e. il n'y a pas de situation possible qui soit représentée comme impossible.
5. Chaque représentation indique une possibilité authentique.

Notons simplement que dans le cas de systèmes diagrammatiques clos sous contraintes, les représentations exhibent la plupart de ces caractéristiques, et en particulier 2, 4 et 5.

Pour conclure cette section nous dirons que les caractéristiques des systèmes de représentations diagrammatiques selon B&E sont essentiellement dérivées de l'existence d'un homomorphisme structurel entre les représentations syntaxiques et la situation représentée. Dans les cas les plus paradigmatiques, ces systèmes possèdent une propriété particulière, la clôture sous contraintes, qui permet au système d'inclure les conséquences logiques des faits constitutifs de la situation sans nécessiter de calcul. Ces systèmes sont donc particulièrement adaptés à la représentation de conjonctions de faits, mais souvent limités en ce qui concerne la représentation de la disjonction[11], et de manière plus générale, de relations du second ordre. Cette représentation est néanmoins possible dans certains cas, comme le montrent les diagrammes de Venn-Peirce définis par S. Shin, mais du point de vue calculatoire, on a alors les mêmes difficultés que les systèmes linguistiques.

## III.2. Spécificité et abstraction limitée

Keith Stenning et Jon Oberlander pensent que ce qui distingue principalement les représentations linguistiques des représentations graphiques est leur pouvoir d'abstraction. Les représentations graphiques se distingueraient par un pouvoir limité d'abstraction, qui augmenterait leur efficacité calculatoire. A l'inverse, les systèmes linguistiques auraient un pouvoir d'abstraction illimité, mais une capacité calculatoire restreinte. Cette dualité serait l'expression d'un compromis entre expressivité et efficacité, inhérent à tout système computationnel, qu'il soit naturel ou artificiel.

Ils identifient cette limitation de la capacité d'abstraction des systèmes de représentation graphiques avec une propriété qu'ils nomment *spécificité* :

> We term this property of graphical systems of representation *specificity* : the demand by a system of representation that information in some class be specified in any interpretable representation. (Stenning et Oberlander, 1995).

---

[11] Il est amusant de noter au passage que les réseaux de neurones ont eux aussi longtemps été accusés de ne pouvoir représenter les disjonctions, et plus généralement les propriétés non linéairement séparables. En effet, le perceptron monocouche ne pouvait pas réaliser cette tâche. Les connexionnistes savaient depuis longtemps qu'avec des réseaux multicouches, ils pouvaient parvenir à représenter de telles classes mais il aura fallu attendre 1987, pour que soient établis des algorithmes d'apprentissage théoriquement fondés.

Cette notion est en réalité inspirée de la notion de *caractère déterminé*. Il y a deux notions de caractère déterminé selon Perry et Macken (1996), l'une qu'ils nomment « *issue determinateness* » et l'autre « *Berkeley determinateness* ». Le *caractère déterminé selon Berkeley* est le fait qu'il n'est pas possible de représenter un objet comme ayant une certaine propriété sans représenter en même temps une valeur déterminée pour cette propriété. Ainsi, je ne peux jamais représenter un triangle quelconque sur une figure mathématique sans représenter finalement un triangle particulier. De même, il n'est pas possible de représenter un objet coloré sur une image sans en préciser la couleur. Par contre, je peux parfaitement dire « cet objet a une couleur intéressante » sans avoir précisé laquelle.

L'autre notion de *caractère déterminé* – que recouvre la spécificité de S&O – est une sorte de contrainte obligeant à rendre explicite des valeurs de propriétés autres que la valeur de la propriété que l'on cherche à représenter pour pouvoir fournir une représentation. Cette notion est une sorte de généralisation de la précédente. A l'extrême, toutes les questions que l'on peut être conduit à se poser à propos d'un objet que l'on cherche à représenter devraient être résolues pour que sa représentation puisse être fournie. Ainsi, dans *Hyperproof*, si je pose un objet sur la grille, je dois en préciser la forme, la taille et la position. Je ne pourrais représenter « A et B ont des tailles différentes » sans préciser « A plus grand que B » ou « B plus grand que A ».

C'est aussi la spécificité qui fonde selon Fred Dretske la distinction analogique/digital :

> I will say that a signal (structure, event, state) carries the information that *s* is *F* in *digital* form if and only if the signal carries no additional information about *s*, no information that is not already nested in *s*'s being *F*. If the signal does carry additional information about *s*, information that is *not* nested in *s*'s being *F*, then I shall say that the signal carries this information in analog form. When a signal carries the information that *s* is *F* in analog form, the signal always carries more specific, more determinate, information about *s* than that it is *F*. Every signal carries information in both analog and digital form. (Dretske 1981, p. 137).

Cette propriété analogique des systèmes de représentation graphiques a des conséquences sur leur efficacité dans le traitement de l'information. Bien qu'isolée des propriétés visuelles de bas niveau, la spécificité des systèmes graphiques relèverait tout de même selon Stenning et Oberlander du niveau de l'implémentation, mais il s'agirait ici de l'implémentation « de la logique elle-même ». Deux systèmes peuvent être logiquement équivalents du point de vue de l'information qu'ils véhiculent, ils ne sont cependant pas pour autant nécessairement *computationnellement* équivalents si leurs capacités calculatoires ne sont pas les mêmes. S&O pensent qu'il y a toujours une équivalence logique entre les inférences exprimables dans des systèmes graphiques et celles qui sont exprimables dans des systèmes linguistiques, et que la différence entre les deux types n'est finalement qu'une différence d'implémentation qui affecte le coût des inférences[12].

---

[12] Cependant, comme il s'agit d'une différence dans l'implémentation de la logique elle-même, les mots *implémentation* et *computationnel* ne sont donc pas à entendre ici avec les sens introduits initialement par David Marr, pour qui le niveau de l'implémentation était le niveau le plus bas (celui du matériel), et le niveau le plus haut, celui des fonctions – le niveau computationnel étant défini comme une sorte de niveau intermédiaire entre les deux (Marr 1982). Ici, le niveau le plus haut est le niveau logique, mais avec un sens un peu inhabituel pour ce mot, puisque des systèmes diagrammatiques, comme des systèmes logiques linguistiques traditionnels, sont alors considérés comme des niveaux computationnels correspondant à des implémentations distinctes d'une même logique plus abstraite.

Ils introduisent trois classes de systèmes de représentation : les MARS (*Minimal Abstraction Representational Systems*), les LARS (*Limited Abstraction Representational Systems*) et les UARS (*Unlimited Abstraction Representational Systems*) – les systèmes de représentation linguistiques faisant bien sûr partie de la dernière catégorie.

Un MARS est un système dans lequel une représentation correspond à un unique modèle du monde sous l'interprétation considérée. Par exemple, la représentation d'une combinaison ordonnée de pions de couleurs au jeu du MasterMind formée d'un quintuplet de lettres symbolisant des couleurs (comme [R R R V B]) permet de définir un modèle. Une table de 0 et de 1, pour reprendre l'exemple de S&O, représentant les valeurs de cinq prédicats unaires pour quatre objets d'un monde W, fournira aussi une représentation d'abstraction minimale si chacune de ses cases ou cellules est occupée par un 0 ou un 1, et rien d'autre – i.e. s'il n'y a pas de cellule vide et pas non plus de cellule occupée par plusieurs symboles.

Les MARS ne permettent l'expression de la disjonction qu'au moyen de plusieurs diagrammes et ne peuvent donc être utilisés que dans des cas où le nombre de prédicats et d'objets est restreint. C'est ce qui se produit dans l'exemple des 5 chaises. La grande limitation de ces systèmes est en effet qu'il faut *n* diagrammes pour représenter *n* modèles. Les MARS n'ont donc aucune capacité d'abstraction, et chaque nouvelle dimension amène avec elle autant de mondes alternatifs que de valeurs spécifiables dans cette dimension.

Mais la plupart des systèmes de représentation diagrammatiques n'imposent pas une telle restriction sur leurs représentations : on peut effectivement augmenter facilement le nombre de modèles capturés par un diagramme d'abstraction minimal en introduisant de nouveaux symboles permettant d'abstraire selon une dimension en autorisant des sous-ensembles de valeurs sur cette dimension. Ainsi, une case vide dans une table de 0 et de 1 peut avoir la signification de représenter un 0 ou un 1. Une table ayant une case vide correspondra alors à deux modèles du monde, une table en comprenant deux, à quatre, etc., et ce, de manière exponentielle. Ainsi, dans l'exemple des chaises, le diagramme du premier cas envisagé recouvrait en réalité les cas 1.1 et 1.2. Mais si les systèmes de ce type permettent apparemment de quantifier massivement sur plusieurs modèles, ils n'en ont pas moins une capacité d'abstraction limitée, car ils ne permettent pas de définir des dépendances arbitrairement complexes entre les différentes dimensions spécifiées. C'est pourquoi S&O les qualifient de LARS. L'abstraction n'est réalisée que sur des modèles qui diffèrent sur les valeurs d'un objet selon une dimension, mais pas sur la nature des rapports et des contraintes qui peuvent exister entre ces différentes dimensions. Seuls des symboles linguistiques, ajoutés à une représentation, peuvent permettre, pensent-ils, de décrire des dépendances arbitrairement fines.

Ainsi, dans l'exemple de la table, on peut imaginer exprimer des dépendances, soit dans la table elle-même, en faisant figurer des équations à la place des symboles uniques dans les cellules (comme on le fait aujourd'hui, avec un tableur), soit à l'extérieur de la représentation de la table, en complétant celle-ci par une série d'assertions. Ainsi, dans la définition de S&O, un système est un *système de représentation d'abstraction illimitée* (noté UARS), s'il exprime des dépendances, soit à l'intérieur d'une représentation, avec des équations ou autres, soit à l'extérieur, avec des assertions sur les « *clés* » définissant la représentation elle-même[13].

---

[13] Nous avions noté une certaine dualité global/local (ou contextuel/non contextuel) opposant les représentations diagrammatiques et les représentations linguistiques pour l'analyse de la clôture sous contraintes, qui semblait accompagner l'opposition explicite/implicite au niveau des données. Cette idée réapparaît dans ces formulations définissant les UARS et les LARS de S&O, et l'on constate à nouveau ici que dans les systèmes linguistiques les relations configurationnelles doivent en effet être explicitées, alors que dans les systèmes diagrammatiques, elles sont souvent implicites. Mais il existe peut-être un autre aspect de cette

Pour capturer la différence intuitive entre les performances calculatoires des LARS et celles des UARS, qui résident selon S&O dans l'expressivité combinée *de la représentation et de ses clés* plus que dans la représentation elle-même, ils décident de qualifier de LARS un système qui maintient « ses représentations simples et ses assertions externes aux clés des représentations ». Ils prétendent alors que la grande majorité des systèmes de représentations diagrammatiques sont des LARS. Les LARS construits sur des MARS par ajout de symboles ont effectivement une capacité d'abstraction limitée. On augmente le nombre de modèles pouvant être appréhendés par un diagramme minimal, mais un diagramme comportant $m$ occurrences d'un nouveau symbole ne peut guère faire mieux que de représenter simultanément un nombre de modèles exponentiel en $m$. Cette abstraction est limitée par la nature des partitions qu'elle permet de réaliser sur l'ensemble des modèles. Pour résumer, les LARS ainsi construits ont un pouvoir d'abstraction limité car ils ne peuvent rendre compte de régularités concernant les *configurations internes* de ces représentations. L'efficacité calculatoire d'un système de représentation étant essentiellement basée sur sa capacité à éliminer de larges branches de cas alternatifs, de tels systèmes ne seront efficaces que si la description initiale des objets est formulée selon les « bonnes » dimensions relativement au problème posé.

La vision de S&O est originale et éclairante car elle s'applique effectivement à de nombreux systèmes diagrammatiques, et elle explique parfaitement leurs propriétés. Nous en avons constaté la pertinence dans une étude sur le raisonnement d'un joueur de MasterMind (Recanati, 2004). Mais nous pensons que l'analyse de S&O n'a pas été poussée suffisamment loin, et qu'elle ne rend pas compte du caractère propre des fonctions sémantiques de ces systèmes. Elle aboutit en effet à une sorte de caractérisation logique qu'ils rapprochent de la hiérarchie de langages de Chomsky en trois groupes : les langages à états finis, les langages indépendants du contexte, et les langages sensibles au contexte. L'analogie est séduisante et certainement pertinente. Mais les langages analysés par Chomsky sont tous constitués de séquences linéaires de symboles arbitraires, et donc linguistiques. Dans la vision de S&O, qui prônent une équivalence logique toujours possible entre les systèmes de représentation graphiques et les systèmes linguistiques, cela n'est guère gênant, mais la notion de spécificité relative à l'abstraction qu'ils ont ainsi mise en relief est peut-être inadéquate pour capturer réellement la distinction linguistique/diagrammatique. Elle ne met en effet en relief que les limitations des représentations graphiques vis-à-vis d'un certain type d'abstraction, appréhendable par les structures linguistiques, sans retenir le fait que les différences profondes qui existent entre les deux types de représentation, notamment le fait que les représentations graphiques soient des représentations bidimensionnelles, pourrait également doter les représentations diagrammatiques de capacités d'abstraction nouvelles, voire supérieures.

C'est en effet relativement à cette notion de dimension géométrique que nous aurions aimé voir émerger une caractérisation des systèmes de représentations diagrammatiques. En effet, les représentations linguistiques sont en quelque sorte des représentations géométriques de dimension 1 (des séquences linéaires de symboles), et les représentations diagrammatiques les plus courantes, des représentations géométriques de dimension 2. Malheureusement, aucun des auteurs que nous mentionnons ici n'a abouti à une caractérisation mettant en relief cette différence intuitive.

### III.3. Localisation et inférences perceptuelles

Jill Larkin et Herbert Simon avaient, en un certain sens, cherché la spécificité des diagrammes dans leur dimension spatiale, car ils ont mis l'accent sur l'utilisation des emplacements pour grouper l'information (Larkin et Simon, 1987). Ils donnent trois raisons qui

---

dualité global/local ayant trait à la relation partie/tout qui pourrait expliquer la non compositionalité et l'irréductibilité des diagrammes aux textes.

permettent aux diagrammes de se révéler supérieurs à des descriptions linguistiques dans la résolution de problèmes :

• Les diagrammes peuvent regrouper ensemble les informations devant être utilisées ensemble, évitant ainsi la recherche d'éléments nécessaires pour réaliser une inférence dans la résolution de problèmes.

• Les diagrammes utilisent typiquement les places pour grouper les informations relatives à un seul élément, évitant d'avoir à comparer des étiquettes symboliques.

• Les diagrammes supportent de manière automatique un grand nombre d'inférences perceptuelles, qui sont extrêmement naturelles aux humains.

Les deux premières raisons expliquent l'efficacité des systèmes diagrammatiques par le fait de regrouper l'information sur un objet unique. Cette propriété serait liée selon Perry et Macken (1996) à la propriété logique de n'utiliser qu'un seul exemplaire d'un symbole pour représenter un objet. Cette propriété, qu'ils nomment *localisation*, disparaît disent-ils, quand on utilise un système de types, et ils observent que les représentations avec des symboles multiples sont omniprésentes dans le langage. Beaucoup de symboles différents du même type et désignant le même objet peuvent être dispersés dans un document, de sorte que l'information concernant cet objet n'est pas « localisée ».

Or, c'est une caractéristique de la perception de rendre accessible l'information sur un individu en inspectant cet individu lui-même, et la localisation donnerait ainsi aux diagrammes et aux images la propriété de fournir des inférences perceptuelles. Regarder un diagramme ou une figure représentant des objets est en effet analogue à regarder ces objets eux-mêmes, et les trois raisons invoquées par L&S n'en seraient finalement qu'une.

Mais cette troisième caractéristique des diagrammes, notée par de nombreux auteurs, est une notion floue relativement à un système formel. Ce qui peut être retenu est que certaines inférences effectuées grâce aux diagrammes seraient peu coûteuses et automatiques, car immédiatement « perçues ». Il s'agit donc d'une propriété liée à l'efficacité computationnelle, mais aussi d'une propriété structurelle qualitative. On peut en cela la rapprocher de la clôture sous contraintes telle que nous l'avons explicitée. Dans un système artificiel linguistique, la contrainte du symbole unique peut en effet être simulée par des artifices d'implantation ou des heuristiques pour réussir à exploiter ce regroupement de propriétés du point de vue de l'efficacité[14], mais ce regroupement de données doit posséder d'autres propriétés pour obtenir des inférences automatiques comme celles provenant du caractère déterminé des représentations.

Notons en outre, même si cela paraît anecdotique, qu'un système diagrammatique artificiel devrait peut-être posséder des procédures d'inspection sur les représentations permettant de vérifier qu'un fait est ou non représenté pour simuler l'existence de ce type d'inférence, et que ces procédures auront alors un coût. Ce n'est que si ces procédures d'inspection sont peu coûteuses qu'un « raisonnement » diagrammatique ne coûterait alors rien en temps calcul. Se pose aussi la question, dans un système artificiel, du *déclenchement* de telles procédures. Ces questions sont aussi des questions d'architecture (non plus de données, mais de programmes). Concernant la modélisation cognitive, la question de l'architecture est d'autant plus importante, que la construction des représentations est à mettre en rapport avec leur mémorisation[15].

---

[14] Comme des tableaux d'indexation dont les accès sont optimisés (hash-code).

[15] Ce qui se passe chez l'homme est parfaitement adapté à ses propres fonctions, et toutes ses capacités sont en quelque sorte intégrées. En particulier, la mémoire, la perception et le langage sont intégrés dans le cerveau, et les ordinateurs ne peuvent guère lutter contre des millions d'années d'évolution (sans compter les années d'apprentissage du bébé qui rendent notre espèce si particulière) ! Les modélisations cognitives que nous pouvons donner ne sont

## III.4. Spécificité, clôture sous contraintes et localisation

Bien qu'apparemment proche, la spécificité de S&O est plus forte et distincte de la clôture sous contraintes. Si l'on doit par exemple représenter sur une droite que B est à gauche de A, et que C est également à gauche de A, on sera obligé de situer C par rapport à B. Mais cette caractéristique n'est pas nécessaire : on peut avoir une représentation diagrammatique où A est devant B et C, avec un angle de vue tel que A masque la position relative de B et C. On peut donc avoir une représentation diagrammatique de cette situation qui n'aurait pas contraint à préciser si B est devant C ou le contraire, comme illustré Figure 9.

C'est également la conclusion de Perry et Macken qui montrent que des représentations qui sont systématiques, contraintes et localisées sont closes sous contraintes mais n'ont pas nécessairement un caractère déterminé. Ils retiennent de leur analyse quatre caractères discriminants des systèmes de représentation : 1) l'iconicité, 2) l'interprétation contrainte et systématique, 3) la localisation et 4) le caractère déterminé (au sens de Berkeley).

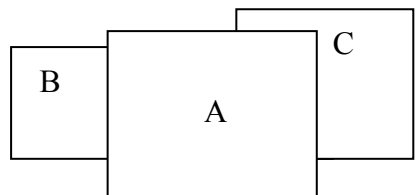

Figure 9. A devant B et C

Le caractère *systématique* précise que l'homomorphisme entre la situation et la représentation n'est pas fixé morceau par morceau, mais de manière globale. Ainsi, lorsqu'une mesure sur une carte indique qu'un cm représente par exemple un km, toutes les mesures de distances sur la carte sont alors interprétables sans qu'il y ait eu besoin de les spécifier une à une[16]. Ce caractère implique que tout le système de relations représente (de manière systématique) un autre système de représentation. Le caractère *contraint* conjoint au caractère *systématique*, assure que l'homomorphisme entre les objets représentés et la situation garantisse le caractère explicite de toutes les conséquences logiques.

L'autre notion qu'avait introduite Stenning et Oberlander et qu'ils pensaient corrélée à la notion de spécificité est la notion de *regimentation* (Stenning et Oberlander, 1991). Pour en donner l'intuition, ils énumèrent des suites de systèmes de représentation, ordonnés selon cette dimension. Ainsi, une quantification abstraite/ un texte désordonné/ un texte ordonné ; ou encore, une table alphabétique des distances entre villes/ la même table avec les villes ordonnées selon leur longitude sur les colonnes et leur latitude sur les lignes/ une carte géographique. La notion est relativement floue, mais en partie corrélée au nombre de modèles qui peuvent valider une représentation, et donc à la notion de spécificité. Le caractère systématique introduit par P&M recouvre en partie cette notion.

Le critère supplémentaire qu'ajoutent P&M pour obtenir la clôture sous contraintes est la *localisation*. Ce caractère rend compte selon eux des observations de Larkin et Simon et ils le rapprochent du critère logique qu'ils appellent « la contrainte de l'exemplaire unique ». Cette

---

donc jamais que de grossières approximations qui ne peuvent que révéler des principes généraux. Cela dit, certaines notions liées à la perception humaine, comme le focus ou la saillance, pourraient être intégrées dans une théorie interprétative et calculatoire, pas nécessairement pour modéliser ce qui se passe chez un humain, mais pour enrichir (en quelque sorte par mimétisme), les possibilités de l'ordinateur.

[16] A noter que ces distances ne concernent que les longueurs des routes, pas leur largeur !

contrainte assure que l'information sur un objet individuel est regroupée en un lieu unique, qui en garantit l'accès rapide et immédiat (et donc, un coût de calcul limité). P&M indiquent qu'il y a bien un aspect qui relie la notion de localisation à la régimentation, et que c'est de cet aspect qu'ils rendent compte par leur catégorisation finale. Mais ils ne discutent pas plus cette question et donnent finalement ainsi une interprétation purement logique à cette dimension (bien qu'ils aient gardé le terme de *localisation*). La localisation est finalement une propriété qui permet de classer des systèmes logiques en quantifiant sur le nombre de symboles utilisés pour représenter les objets, mais cette caractéristique, comme celle de S&O, ne rend pas compte des caractères géométriques ou iconiques des représentations utilisées par les systèmes en question. En particulier, ils ne rendent pas compte des possibilités particulières de *partitionnement* des dimensions de spécificité qui sont ainsi offertes[17] par des représentations géométriques de dimension 2.

Avec ces quatre propriétés, P&M distinguent finalement 5 catégories de représentations allant des textes aux images : les textes, les schémas graphiques (*charts*), les diagrammes (*diagrams*), les cartes (*maps*) et les images (*pictures*). Cette catégorisation générale de P&M est illustrée Figure 10.

|  | Iconicité | Interprétation contrainte et systématique | Caractère déterminé de Berkeley | Localisation |
|---|---|---|---|---|
| Textes | non | non | non | non |
| Graphiques | non | non | non | partielle |
| Diagrammes | partielle | oui | non | totale |
| Cartes | partielle | oui | oui | totale |
| Images | totale | oui | non | totale |

Figure 10. Les cinq types de représentations selon Perry et Macken

## III.5. Iconicité et sens profondément enraciné

Le caractère *d'iconicité* qui caractérise les diagrammes ne serait donc pas lié selon P&M au caractère déterminé des représentations, mais plutôt à la localisation. L'iconicité avait été particulièrement discutée par Macken, Perry, et Cathy Hass (Macken et alii, 1993). Selon eux, ce caractère graphique et spatial dote les représentations d'un *sens profondément enraciné* (ou

---

[17] Les lieux (ou places) en dimension 2 permettent en effet de regrouper ensemble des traits différents, comme la couleur, la texture ou autres, soit dans une même région de l'espace, comme c'est le cas pour des objets physiques, soit en des régions séparées (on fait alors un regroupement sur la base que ces lieux partagent les mêmes traits). On peut alors, en observant les formes obtenues, « visualiser » des interférences entre plusieurs dimensions de traits qui « se croisent » pour reprendre la terminologie de Jean Nicod :

"Ce qui importe, c'est l'existence d'une ressemblance locale et d'une ressemblance qualitative qui divergent, c'est la présence de deux réseaux de ressemblances qui se croisent, classant les données de deux manières différentes. Le toucher et la vue possèdent seuls cette richesse de structure. C'est pourquoi ils donnent des géométries particulièrement intéressantes" (Nicod, 1962).

signification richement fondée : RGM – *Richly Grounded Meaning*). Une signification profondément enracinée est une signification qui n'est pas arbitraire. Cela peut venir de divers facteurs reliant la forme du symbole à sa signification. Un signe iconique peut avoir un sens facile à déduire (RIM – *Readily Inferable Meaning*), un sens facile à mémoriser (ERM – *Easily Remembered Meaning*), ou encore, un sens modifiable de manière interne (IMM – *Internally Modifiable Meaning*).

Les panneaux de signalisation routière fournissent de nombreux exemples de sens facile à déduire et à mémoriser (par exemple, les panneaux signalisant des virages). On trouve aussi, dans les partitions musicales, des exemples de signes dont le sens est modifiable de manière interne, de par le caractère analogique du signe utilisé (par exemple, les notations de crescendo situées sous la portée).

Perry et Macken contribuaient dans cet article au projet *Archimedes* du CSLI, dont l'objet était d'améliorer l'accès à l'information des personnes handicapées. L'idée centrale du projet était que les handicapés se trouvaient gênés par les systèmes informatiques ne rendant accessible l'information que sous une forme, car cette forme pouvait nécessiter des capacités perceptives ou motrices particulières, alors que des interfaces plus générales, ou peut-être plus spécifiquement adaptées, pouvaient être définies. En effet, le développement des interfaces graphiques a posé d'importants problèmes aux aveugles qui avaient l'habitude de travailler sous interface textuelle classique (de type DOS) à l'aide de dispositifs convertissant automatiquement les mots en voix.

C'est dans le cadre de cette étude qu'ils se sont intéressés aux systèmes de communication hétérogènes et notamment au langage des sourds, en tant qu'exemple illustrant les caractéristiques de tels systèmes. Le langage des signes américain (ASL – *American Sign Language*) a été pour eux un objet d'étude particulièrement intéressant parce que c'est un langage qui a toutes les caractéristiques et la même puissance d'expression que n'importe quelle langue naturelle[18], et cependant, du fait que c'est un langage de gestes, il offre plus d'opportunités qu'un langage vocalisé d'exhiber des sens profondément enracinés.

Comme on l'a rappelé dans l'introduction, bien des signes peuvent être iconiques au départ, puis se trouver ensuite stylisés, jusqu'à perdre leur propriété d'avoir un sens facile à mémoriser ou à induire. L'iconicité des éléments lexicaux confère donc un caractère iconique à ASL, mais de manière partielle, et ce sont plutôt les possibilités d'enracinement dites du « second-ordre » qui ont intéressé P&M. Les locuteurs habiles peuvent en effet utiliser des possibilités spatiales pour modifier créativement le sens. C'est ce qui correspond à la propriété d'enracinement du sens qu'ils appellent IMM (sens intérieurement modifiable)[19].

Dans l'analyse de la manière dont ASL utilise l'espace, ils retrouvent trois des cinq catégories de représentation qu'ils ont isolées Figure 10. Ils distinguent en effet dans ASL trois modes (ou états) qu'ils nomment, le *Texte*, l'*Organisation par Région*, et la *Scène*, (qui correspondent aux lignes 1,2 et 4 de la table), et il ressort de leur analyse que ASL est un système de communication *hétérogène*, articulant ces trois modes de représentation[20].

---

[18] Cependant, selon P&M, il y a aujourd'hui une importante littérature où l'emphase est mise sur les similarités plutôt que sur les différences, entre ASL et l'anglais.

[19] Les langages qui ne possèdent pas de telles ressources iconiques trouveront d'autres moyens, comme par exemple l'utilisation d'adverbes, pour modifier le sens.

[20] Le mode Texte correspond à la première ligne et n'a aucune des quatre caractéristiques des représentations diagrammatiques. L'Organisation par Région correspond à la seconde, et le mode Scénique est analogue à la catégorie des cartes (ou plans) car il utilise l'espace pour y construire des sortes de schémas.

Nous pensons comme Perry et Macken que le langage naturel met aussi en jeu un système de représentations hétérogènes (SRH), même si on le catégorise a priori comme linguistique. Le paradigme des systèmes linguistiques, en tant qu'opposés aux systèmes diagrammatiques, est donc pour nous plutôt incarné par les systèmes de représentations logiques, que par un texte en langage naturel, dans la mesure où on s'intéresse aux représentations cognitives mises en jeu pour sa compréhension - ces représentations n'ayant a priori aucune raison d'être purement linguistiques si elles sont, comme nous le croyons, ancrées dans la perception.

Pour en revenir aux représentations diagrammatiques, l'analyse du caractère iconique des symboles de P&M met bien en évidence ses conséquences pour le sens, i.e., le fait que le sens soit alors profondément enraciné dans le système de représentation et dote l'ensemble de propriétés nouvelles (facilité d'interprétation, de mémorisation ou d'émergence de nouveaux sens). Mais les considérations développées sont trop générales pour identifier comment ces conséquences sont liées à l'espace et à la structure spatiale des représentations. Avant de conclure, nous allons ajouter deux remarques personnelles concernant l'enracinement des symboles iconiques.

### III.5.1. Iconicité et homomorphisme syntaxique

Sous cet aspect, nous pensons que la principale distinction entre les systèmes de représentation « linguistiques » (ou « symboliques ») et les systèmes de représentation « analogiques » (dont feraient partie les diagrammes) peut être caractérisée en termes de puissance du métalangage requis pour donner la sémantique du système : dans le cas des notations analogiques, le métalangage nécessite une référence à des propriétés syntaxiques du langage-objet, et dans le cas des notations symboliques classiques, cela n'est pas nécessaire.

L'exemple des trois figures alignées que nous avons considéré dans la partie III.1. va nous permettre d'illustrer ce point. Soient les deux notations (I) et (II) suivantes, exprimant la même information, la première étant de type symbolique, la seconde de type analogique. La différence, minimale, entre (I) et (II), semble cependant capter une différence essentielle entre les deux types de représentation :

(I)  à-gauche-de(a, b) & à-gauche-de(b, c)

(II)  aligné([a, b, c])

La syntaxe de la première notation spécifie que « à-gauche-de » est un prédicat à deux arguments. Dans ce type de notation, les arguments du prédicat sont en nombre fixe et ordonnés, mais n'ont pas de structure interne pertinente. L'ordre mis entre les arguments provient donc uniquement de la sémantique associée au symbole « à-gauche-de ».

La syntaxe de la deuxième notation spécifie que « aligné » est un prédicat à un argument, mais que cet argument est structuré : c'est une liste ordonnée (qui pourrait d'ailleurs comprendre en réalité un nombre quelconque de termes). La différence entre (I) et (II) est une complexité syntaxique additionnelle de l'argument du prédicat dans (II) puisque l'argument du prédicat est une liste entre crochets. Dans le cas de « à-gauche-de », on dira que à-gauche-de($\alpha$, $\beta$) est vrai *si et seulement si* le référent de $\alpha$ se trouve à gauche du référent de $\beta$. Du point de vue dénotationnel, on a une équation traduisant le fait que le sens de la fonction est fonction du sens de ses arguments :

[[ à-gauche-de( $\alpha$ , $\beta$) ]]= [[à-gauche-de]] ( [[ $\alpha$ ]] , [[ $\beta$ ]] )

Mais avec « aligné », la situation est différente, car ce prédicat *ne peut pas* être décrit comme fonction à *n* arguments, mais seulement comme fonction d'un argument *structuré* (ici d'arité variable) : aligné([a, b, ... n]) est vrai *si et seulement si*, pour chaque paire ($\alpha$, $\beta$) de symboles de la liste [a, b, ... n] tel que $\alpha$ précède $\beta$ dans la liste, le référent de $\alpha$ se trouve à gauche du référent de $\beta$. La formulation de la règle sémantique pour « aligné » est donc plus complexe que celle de « à-gauche-de » puisqu'elle doit mentionner non seulement des symboles du langage-objet mais aussi *leur position dans une structure de donnée syntaxique*.

Le sens de cette fonction ne pourra être fonction du sens de son argument que si on le maintient tel quel dans une équation du type :

[[ aligné ([a, b, ... n]) ]]  =  [[ aligné ]] ( [[  [a, b, ... n] ]] )

La fonction interprétative ne peut donc plus être définie en composant des termes fonctionnels et des termes simples, elle doit aussi assigner un sens à des *configurations* de termes, comme celle définie ici par les crochets simples : [a,b, …,n]. On ne peut donc pas décrire le sens de « aligné » directement en fonction du sens des éléments figurant en argument dans la liste[21]. On doit avoir *au niveau du métalangage de description sémantique* les *mêmes possibilités de structuration syntaxique de données* que celles qui figuraient dans le langage de représentation, et l'on doit aussi leur assigner un sens. En d'autres termes, la sémantique d'une notation analogique serait définie non seulement sur les symboles primitifs de la représentation, mais aussi sur des structures de données internes à la représentation, utilisées précisément pour leurs aspects configurationnels dans cette représentation.

Cependant, il n'est bien entendu pas nécessaire que toute nuance dans le système de représentation s'exprime dans le système d'interprétation. En effet, tous les traits, et en particulier tous les traits iconiques, ne sont pas interprétés dans une représentation picturale. On utilise parfois des figures volontairement fausses en géométrie, mais sur lesquelles, on effectue des raisonnements justes, en marquant les quantités égales par des traits semblables, sur les longueurs de segments ou des angles : là, seuls ces traits semblables auront une signification, pas les segments accidentellement de même longueur.

De même la structuration du symbole n'est pas fondamentalement nécessaire pour que soit exploité un caractère iconique des symboles. Ainsi, avec l'ajout d'un symbole unique (et donc non structuré) à un MARS comme proposent de le faire S&O, on peut obtenir un LARS qui reflète des relations structurelles non explicites. On peut par exemple imaginer avoir une table dont les valeurs seraient prises parmi deux symboles, 0 et 1, mais pouvant figurant en caractère gras ou non gras. On aurait donc finalement quatre valeurs (0, **0**, 1 et **1**) mais les interprétations sémantiques pourront utiliser le fait que les formes 0 et **0,** ou 1 et **1**, désigne une même valeur. Dans un tel système, où la forme des symboles n'est plus arbitraire, on peut effectuer d'autres abstractions que celles naturellement induites par le nombre de dimension de spécificité, et abstraire ainsi sur des combinaisons « croisées » de dimensions.

### III.5.2. Iconicité et calcul

Chez tous les auteurs que nous avons rencontrés, l'attention sur l'aspect iconique des symboles n'a porté que sur l'homomorphisme entre monde représentant et monde représenté. Cela se traduit sur le schéma de la Figure 11, par des observations concernant le rapport des symboles figurant sur la première ligne, qui représente cette correspondance (c'est la correspondance classique entre syntaxe et sémantique). C'est relativement à cet axe qu'ont été décrites les caractéristiques iconiques des systèmes de représentation diagrammatiques relevées

---

[21] En effet, si l'on essaye d'exprimer le sens de cette configuration en fonction des termes simples qui la constitue, on ne pourrait qu'écrire

[[ [a, b, ... n] ]] = [ [[ a ]], [[ b ]], .... , [[ n ]] ] ,  d'où l'équation sémantique :

[[ aligné ]] ( [[  [a, b, ... n] ]] ) = [[ aligné ]] ( [ [[ a ]], [[ b ]], .... , [[ n ]] ] )

Mais on n'aura pas réussi ainsi à se débarrasser des crochets simples (donc de syntaxe) dans la partie droite de l'équation sémantique.

par les auteurs que nous avons mentionnés, et c'est également relativement à cet axe que se situe la remarque précédente.

Mais pour être implanté sur une machine, cet homomorphisme sera traduit en un algorithme qui calculera à partir de représentations internes, d'autres formes de représentations internes, explicites et externalisables. Relativement à ce nouvel axe (que nous appelons volontiers *axe du calcul*), l'iconicité des symboles peut aussi faire apparaître des sens « profondément enracinés ».

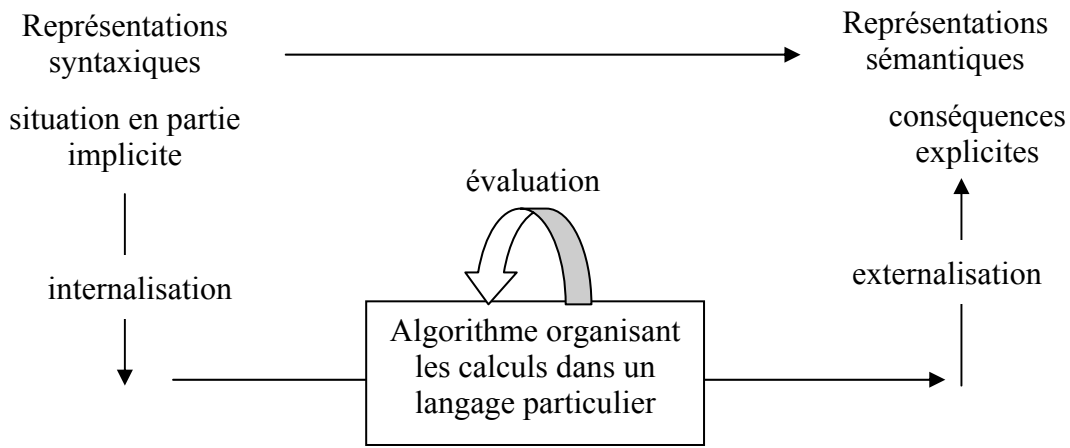

Figure 11. L'axe du calcul

Il existe en effet une nouvelle variété de moyens d'établir des correspondances entre les deux types de représentations. En effet, des liens non arbitraires peuvent lier la syntaxe du langage de programmation utilisé aux représentations (lors de l'opération *d'internalisation*, de celle *d'externalisation*, ou même lors du calcul[22]).

A titre d'exemples, certains langages de programmation des années 80 avaient des caractéristiques originales, comme des capacités de *pattern-matching*, qui augmentaient leurs possibilités d'interprétation fonctionnelle des structures syntaxiques (cf. le langage d'acteurs PLASMA). Dans les langages de programmation dits « réflexifs », un objet peut être considéré contextuellement tantôt comme un programme pouvant s'exécuter, tantôt comme une donnée sémantiquement interprétable[23], et l'on peut manipuler la représentation du programme lui-même pour la modifier en cours d'interprétation. La distinction programme/donnée est dans ce cas remise en cause[24] et une plus grande richesse de relations entre monde représentant et monde représenté est virtuellement possible[25]. Mais nous n'entrerons pas dans ces questions sémantiques, car elles sont trop complexes pour être discutées ici, et relèvent du domaine de la programmation et de sa théorie.

---

[22] De surcroît, le processus d'évaluation ne s'effectuant pas nécessairement en un coup, un symbole peut avoir une forme non arbitraire à certains moments du processus d'évaluation, et pas à d'autres, et l'établissement de la correspondance entre les deux types de représentations (internes ou externes) peut donc utiliser un caractère d'iconicité ancré dans le langage de calcul lui-même (comme le caractère gras, pour introduire une notion de focus). La notion d'évaluation n'est d'ailleurs pas aussi simple qu'elle paraît l'être sur ce schéma. Dans le cas de langage de programmation dits réflexifs, des opérations « d'internalisation » ou d'« externalisation » peuvent se produire à différents moments. Une évaluation peut être partielle, ou temporisée, suspendue ou relancée selon le contexte.

[23] C'est le cas avec le langage lisp, et notamment avec les langages d'assemblage, car la machine, au plus bas niveau, ne fait pas la distinction entre programmes et données.

## III.6. Résumé

La plupart des caractéristiques des systèmes diagrammatiques relevées par Barwise et Etchemendy semblent fondées sur l'existence d'un homomorphisme structurel entre les représentations syntaxiques et la situation du monde représenté. Dans les cas les plus paradigmatiques, le système possède une propriété de clôture qui fait que toute conséquence des faits initiaux est incluse dans la représentation sous l'effet de contraintes structurelles et ne nécessite pas de calcul. Cette situation est en contraste brutal avec celle des systèmes logico-linguistiques, où la plus triviale conséquence doit être inférée explicitement. Mais, si ces systèmes ont des performances impressionnantes dans des tâches de simulation où il s'agit de

---

[24] La notion est de toute façon remise en cause aujourd'hui, mais pour d'autres raisons. Le développement récent des interfaces graphiques et des réseaux d'ordinateurs a en effet modifié la donne relativement à la notion d'entrée/sortie qui avait permis de définir la sémantique des langages de programmation et des programmes en s'appuyant sur un modèle mathématique et la compositionalité fonctionnelle. Actuellement, les outils sémantiques dont nous aurions besoin pour décrire la sémantique d'un programme utilisant une bibliothèque graphique comme celle d'un système de fenêtrages n'existent tout simplement pas, car ces bibliothèques ne sont plus des bibliothèques de fonctions basées sur un modèle mathématique, mais des bibliothèques de programmes complexes dont la sémantique n'est pas nécessairement bien définie.

Au début de l'informatique, les ordinateurs ne manipulaient en effet que des données linguistiques, puisque leurs entrées étaient constituées de la frappe de touches au clavier, ou de fichiers contenant des séquences linéaires de caractères alpha-numériques. Leurs sorties avaient ce même caractère textuel, car elles ne pouvaient se produire que sur des imprimantes ou des écrans aux possibilités très limitées. Le développement des interfaces graphiques et des réseaux a introduit plusieurs éléments nouveaux, modifiant radicalement ce cadre : (1) il y a maintenant d'autres sources d'entrées (par exemple la souris), et d'autres sources de sorties (l'écran graphique, ou le son) ; (2) ces entrées/sorties de nature différentes sont néanmoins intrinsèquement reliées entre elles dans le système (comme le sont actuellement la souris et l'écran graphique) ; (3) A cette complexité additionnelle s'est ajouté celle du partage des entrées/sorties par plusieurs programmes pouvant se dérouler simultanément et communiquant entre eux (par l'intermédiaire du serveur d'entrées/sorties).

La sémantique d'un programme ne peut donc plus être définie indépendamment de tout système, en ne considérant que des niveaux de langages symboliques empilés ne manipulant finalement que des chaînes. Il faut établir des niveaux d'abstractions intermédiaires souvent complexes et eux-mêmes susceptibles d'articulations, établissant comment les entrées/sorties sont reliées entre elles et comment s'effectue le partage entre les différents programmes y ayant accès. Ainsi, le sens d'un programme *ne peut plus être complètement indépendant du sens des autres programmes qui s'exécutent au même moment que lui*. Un exemple simple : l'exécution d'un même code de programme qui affiche une fenêtre à large bord est dépendante de la présence (ou de l'absence) d'un gestionnaire de fenêtres, car ce dernier peut contrôler l'affichage des bords des fenêtres créées par d'autres programmes.

[25] D'autres possibilités comme celles en lisp, de définir des macros, ou de suspendre l'évaluation d'un terme, permettent également d'abstraire sur des procédures, et introduisent de nouvelles possibilités d'abstraction.

représenter des conjonctions de faits, ils sont généralement inadaptés à la représentation de disjonctions et de relations du second ordre.

Pour Stenning et Oberlander, les représentations linguistiques et les représentations graphiques se distinguent principalement par une capacité d'abstraction limitée qui contraint l'information à être explicitée, augmentant ainsi leur efficacité calculatoire. Leur conclusion est que ce qui caractérise les systèmes de représentations diagrammatiques n'est finalement pas lié à la nature des représentations, mais plutôt à l'expressivité du métalangage qui les utilise. Nous les rejoignons sur cette question du métalangage, mais seulement en partie. Nous pensons en effet que la principale distinction entre les systèmes de représentation « linguistiques » et les systèmes de représentation « analogiques » (dont feraient partie les diagrammes) peut être caractérisée du point de vue formel en termes de puissance du métalangage requis pour en donner la sémantique : dans le cas des notations analogiques, le métalangage nécessite une référence à des propriétés syntaxiques du langage-objet, et dans le cas des notations symboliques classiques, cela n'est pas nécessaire.

Perry et Macken ont discuté les caractéristiques isolées par les auteurs précédents et souligné l'importance du caractère iconique. Ils ont mis en évidence l'apparente corrélation de l'iconicité et de la localisation (ou contrainte de l'exemplaire unique). Mais cette dernière propriété s'avère être comme les autres une propriété formelle indépendante de l'espace et notre conclusion est finalement que la distinction entre textes et diagrammes n'est encore qu'imparfaitement élucidée[26]. Les caractéristiques relevées par tous ces auteurs sont en effet des caractéristiques dans lesquelles l'iconicité et le caractère spatial des symboles restent insuffisamment analysés.

## IV. PERSPECTIVES COGNITIVES ET COMPUTATIONNELLES DES SRH

L'étude précédente a mis en évidence l'intérêt des systèmes inférentiels diagrammatiques, mais l'on pourrait s'interroger sur la pertinence de tels outils – l'homomorphie syntaxique risquant de ne se manifester que dans très peu de cas. L'homomorphie syntaxique est en effet une condition très exigeante si l'on souhaite modéliser tous les aspects d'une situation par un diagramme. Mais dans le cadre de systèmes hétérogènes, l'architecture générale n'est plus la même, puisqu'on articule un module linguistique avec d'autres modules, graphiques ou semi-graphiques[27]. On peut alors bénéficier de plusieurs homomorphismes locaux, permettant de regrouper des aspects liés du modèle. C'est pourquoi les représentations diagrammatiques ont en réalité un vaste champ d'application, mais dans un cadre particulier : celui des systèmes de représentations hybrides ou hétérogènes.

### IV.1. Perspectives computationnelles

Les systèmes de représentation hétérogènes présentent l'énorme avantage sur les autres systèmes, qu'ils peuvent bénéficier des propriétés antinomiques des représentations logiques et analogiques. En particulier, les questions de complexité concernant les calculs se posent différemment dans un SRH. Le paradoxe est qu'une démonstration dont le coût peut être borné théoriquement dans tout système, et donc en particulier dans deux systèmes différents, pourrait avoir un coût inférieur au minimum des deux dans un « système hybride » constitué de leur regroupement, pour autant que l'ensemble puisse exploiter les aspects optimaux de l'un et l'autre. C'est ce que montre intuitivement le schéma de la Figure 12, où les démonstrations minimales dans S1 et S2 peuvent se combiner avantageusement dans un système hybride pour établir un même fait en y produisant en quelque sorte une démonstration diagonale.

---

[26] Une piste de recherche pour éclairer le caractère iconique serait d'analyser sur différents exemples l'ancrage des symboles dans l'homomorphisme syntaxique et dans le calcul.

[27] Il n'y a d'ailleurs que très peu de schémas ou de diagrammes qui ne soient incrustés d'éléments textuels faisant implicitement référence à un modèle linguistique.

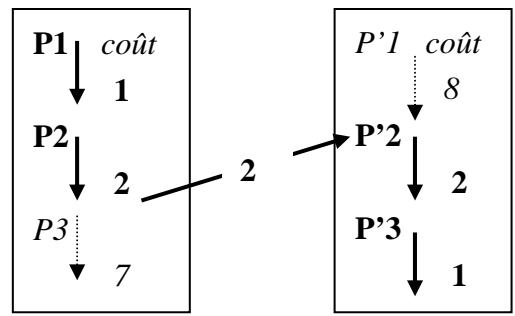

Coût(S1): 1+2+7 = 10   Coût(S2): 8+2+1 = 11
Coût(S1+S2): 1+2+2+2+1 = 8

Figure 12. Coût possiblement inférieur d'un calcul diagonal

La remarque de B&E sur la non nécessité de définir un langage global articulant l'ensemble prend du poids dans cette perspective : on n'a pas besoin de définir un métalangage « plus puissant » ou « aussi puissant » que celui nécessaire à l'ensemble du système pour en fonder la sémantique, et il peut donc en découler une diminution de la complexité effective.

Ainsi, la complexité d'un calcul dans un système hétérogène pourrait se révéler *inférieure* à celle d'une fonction sémantiquement équivalente dans un système symbolique, car le simple fait que les deux systèmes *dénotent les mêmes entités du monde* garantit la cohérence du système global, sous réserve, bien entendu, que les correspondances locales entre les deux systèmes soient fondées. Mais l'important ici est de comprendre que les correspondances qui permettent le transfert d'information d'un système à l'autre peuvent rester partielles et trouver l'essentiel de leur légitimité au niveau dénotationnel.

## IV.2. Perspectives en sémantique cognitive

On a vu finalement que l'on retrouvait sous la dualité linguistique/diagrammatique, la dualité classique logique/analogique et le dilemme bien connu des informaticiens entre calculer et représenter. C'est cette remarque qui donne selon nous tout son intérêt aux systèmes de représentation hétérogènes pour la modélisation cognitive. Nous pensons en effet que :

(1) Le cerveau représente (et mémorise des représentations) plus qu'il ne calcule.

(2) L'intelligence réside dans l'utilisation dynamique de correspondances simples entre sous-systèmes de représentation interconnectables.

Ces hypothèses sont renforcées par les résultats des réseaux de neurones, mais également par les théories cognitives de la communication humaine, comme la théorie de la pertinence de D. Sperber et D. Wilson. Selon eux, une caractéristique de l'interprétation pragmatique des énoncés serait en effet que l'interprétation la plus pertinente, au sens où elle apporte les informations nouvelles les plus intéressantes par rapport au contexte, serait obtenue en maximisant son rapport « intérêt/coût», i.e. le rapport entre le coût de son calcul dans le contexte et l'importance de l'effet produit (Sperber et Wilson, 1986). Mais tournée différemment, cette propriété peut être vue comme une tendance plus générale du cerveau à produire directement les informations pertinentes, en utilisant les structures de données disponibles en mémoire de telle sorte que ce qui est finalement le moins coûteux du point de vue calculatoire coïncide miraculeusement avec l'interprétation pragmatique la plus pertinente.

En effet, ce que nous montrent les systèmes de représentations clos sous contraintes, c'est que pour la mise en place d'une telle architecture, le système n'a pas nécessairement besoin de *calculer ce qui maximise le rapport* pour faire apparaître les interprétations pertinentes, il peut aussi bien *représenter* la situation *de manière telle* que ce calcul soit inexistant, car seules les interprétations les plus pertinentes se trouvent, par construction, représentées.

Nous pensons, comme Barwise et Etchemendy, que les formalismes logiques sont sous bien des aspects fondamentalement inadaptés à la description des capacités inférentielles humaines. C'est le cas, en ce qui concerne les relations temporelles, spatiales et causales, et sans doute de manière générale, concernant les principales catégories et relations que nous utilisons pour combiner nos représentations et nous situer dans le monde qui nous entoure. Cette thèse est renforcée pour le temps et l'espace par l'abondance des schémas que l'on trouve dans la plupart des travaux de sémantique contemporaine[28]. On trouve ce même intérêt pour les diagrammes dans tous les travaux en sémantique des événements, même lorsqu'ils se situent dans une vision chomskienne du langage et/ou lorsqu'ils s'intègrent à des courants de sémantique formelle.

En effet, les travaux sur le temps et l'aspect en langue naturelle ont donné lieu à de nombreux modèles de description basés sur des représentations diagrammatiques (par exemple : Guillaume 1929, Reichenbach 1947, Smith 1991, Hornstein 1993, etc.). Bien que le fait que ces théories utilisent des diagrammes pour modéliser les représentations temporelles ne soit pas en soi une preuve de leur existence dans des représentations mentales, c'est néanmoins un indice, et si ces diagrammes permettent d'expliquer autant les failles que l'aisance des inférences humaines, leur intérêt s'en trouve considérablement renforcé.

L'idée d'utiliser des représentations analogiques pour modéliser les représentations cognitives a réémergé en I.A. au début des années 90. Pour preuve, cet extrait d'une conférence prononcée en décembre 1990 par Daniel Kayser à la Société Française de Psychologie :

> « L'I.A. devrait se pencher beaucoup plus sérieusement qu'elle ne l'a fait jusqu'ici sur l'idée de représentations "partiellement analogiques" ; c'est déjà implicitement cette idée qui a contribué au succès des réseaux sémantiques, dont la plupart sont équivalents à une représentation logique […] et qui possèdent en outre des propriétés proches de représentations analogiques (on peut avoir envie d'y définir des notions de propagation, de proximité, de recherche de motifs, notions qui n'effleureraient pas l'esprit dans la formulation logique « équivalente »). […] C'est probablement le même genre d'idée qui devrait être exploité dans un domaine certainement promis à une rapide évolution : la représentation cognitive des relations spatio-temporelles (parmi les apports les plus récents, on mentionnera les travaux effectués à Toulouse autour de Mario et d'Andrée Borillo, p.ex. [Vieu,1991]). Une représentation « réaliste » (le temps mesuré comme un nombre réel, l'espace repéré en coordonnées cartésiennes ou polaires) n'a manifestement pas les propriétés expliquant la facilité – et les erreurs – des opérations mentales que nous effectuons en permanence. » (Kayser, 1992).

Quelques travaux à cheval entre l'intelligence artificielle et la linguistique ont en effet montré qu'il était possible d'élaborer des programmes basés sur des structures de données particulières (associées aux mots ou aux marques syntaxiques) permettant d'expliquer des inférences temporelles subtiles (cf. Zablit 1991, Vieu 1991, Gosselin 1996). Ce type de travaux est à la fois encourageant et prometteur, même si les systèmes proposés sont parcellaires et comportent un certain degré d'arbitraire[29]. Ils ont le mérite d'avoir montré que les questions qui se posent aujourd'hui en sémantique sont des questions d'architecture[30]. Nous pensons comme D. Kayser que ce type d'approche, basé sur l'existence de modules de représentations spécifiques, aux capacités représentationnelles limitées mais bien adaptées au domaine, est fondé s'il permet de rendre compte des capacités comme des limitations des inférences humaines.

---

[28] On peut citer comme particulièrement représentatif le courant de sémantique cognitive qui s'est développé sur la côte ouest des Etats-Unis, en réaction à la vision chomskienne du langage prônant l'autonomie de la syntaxe. Il n'y a qu'à feuilleter Fauconnier 1997, Langacker 1988 ou Vandeloise 1986, pour se rendre compte de l'importance des diagrammes et des schémas dans l'élaboration de ces théories.

## IV.3. Modélisation cognitive du raisonnement

Un autre domaine privilégié d'application des SRH est celui de la modélisation cognitive du raisonnement. Il existe de nombreux programmes d'I.A. utilisant des diagrammes pour la résolution de problème, en particulier dans les courants issus de la physique qualitative, mais il y a eu peu d'études sur le raisonnement utilisant des SRH. Barwise et Etchemendy ont donné les premiers un exemple de raisonnement hybride avec *Hyperproof*, mais l'imbrication des parties linguistiques et diagrammatiques y reste simple et repose sur les interventions de l'utilisateur du logiciel[31] (il n'y a donc pas de modélisation du raisonnement à proprement parler). Le courant initié par B&E a donné lieu à de nombreuses d'études psychologiques sur les diagrammes d'Euler et de Venn, mais ces études se sont attachées à l'analyse de systèmes purement diagrammatiques. Très peu d'analyses ont finalement portées sur d'autres exemples, bien que les preuves mathématiques ou les jeux de récréation mathématiques proposés par les journaux abondent d'exemples de raisonnements figuratifs très divers, et beaucoup plus hybrides.

Pour contribuer à renouveler le genre, nous présentons dans un rapport le résultat d'une étude sur le raisonnement d'un joueur de MasterMind (Recanati, 2004) mettant en évidence l'utilisation de diagrammes et de divers « modèles mentaux » à la Johnson-Laird (Johnson-Laird, 1983). Les raisonnements que nous avons observés ici sont fondamentalement hybrides et utilisent toutes sortes de schémas. Le jeu du MasterMind se prête en effet à la modélisation du raisonnement, car il force le joueur à raisonner logiquement. Par ailleurs, la géométrie du problème incite à élaborer des représentations diagrammatiques. La représentation iconique de la grille du jeu est utilisée non seulement comme mémoire et support géométrique des preuves, mais également pour simplifier la mise en œuvre du raisonnement général et les retours en arrière. Nous avons constaté que la stratégie des joueurs expérimentés pouvait s'expliquer par le cheminement sur une arborescence de modèles mentaux emboîtés analogues aux LARS de S&O permettant ainsi de converger très rapidement vers le modèle solution.

## V. CONCLUSION

Nous avons cherché dans cet article à raviver l'intérêt théorique pour l'utilisation de schémas et de diagrammes dans les systèmes inférentiels de représentation de connaissances.

---

[29] Un certain nombre de travaux en linguistique et en anthropologie cognitive laisse néanmoins espérer qu'on puisse peut-être mettre en évidence des primitives sémantiques constitutives des représentations humaines du temps et de l'espace.

[30] Nous avions développé ces idées avec R. Carter et P. Zablit en 1993 (Carter et alii, 1993) en proposant une architecture modulaire pour représenter les situations, dans laquelle le caractère diagrammatique des capacités de représentation des modules (pour nous le temps ou l'espace) permettait d'obtenir sans calcul toutes sortes d'inférences. Pour la représentation du temps par exemple, une droite orientée sur laquelle on place les événements dénotés par les verbes permet d'obtenir des inférences semblables à celles de l'exemple des trois figures alignées.

[31] Eric Hammer a fourni également, sur la base des travaux de S. Shin, un exemple de démonstration hétérogène, mais les parties linguistiques sont isolées au début et à la fin de sa démonstration (Hammer 1994).

Après une mise en perspective historique, nous avons réaffirmé, sur les traces de Barwise et Etchemendy, que le préjugé des logiciens et des mathématiciens contre les diagrammes, « simple support pour l'intuition pouvant conduire à des erreurs de raisonnement », n'est en réalité pas fondé. Les diagrammes peuvent en effet être vus comme des objets syntaxiques structurés, et se prêter à des raisonnements logiques dans une perspective formelle. Dire cela ne revient cependant pas à dire qu'un système diagrammatique peut être ramené à un système logico-linguistique. Cela signifie simplement que les diagrammes ne sont pas impropres à la formalisation et qu'ils permettent, tout autant que des symboles linguistiques, des raisonnements logiquement fondés.

Nous avons ensuite cherché à faire le point sur ce qui distingue les systèmes inférentiels diagrammatiques des systèmes linguistiques en parcourant la littérature du domaine qui s'est portée sur cette question dans les années 90. La plupart des caractéristiques des systèmes diagrammatiques relevées par ces auteurs est fondée sur l'existence d'un homomorphisme structurel entre les représentations syntaxiques et la situation du monde représenté. Dans les cas les plus paradigmatiques, toute conséquence des faits initiaux est incluse *de facto* dans la représentation et ne nécessite pas de calcul. Il ne s'agit plus ici de calculer mais simplement de représenter, et une qualité importante des systèmes diagrammatiques, outre leurs performances calculatoires, est alors le caractère déterminé de leurs représentations. Les représentations graphiques ont en effet cette faculté de pouvoir faire émerger au sein même de la représentation des sens cachés, voire nouveaux. Mais cette propriété, liée au caractère iconique des représentations, n'a pas été encore bien analysée. Reste en particulier à comprendre comment intervient l'espace, absent des caractéristiques analysées. Les recherches sur les diagrammes doivent donc être poursuivies, et pour élucider la distinction texte/diagramme, il faudra peut-être mobiliser d'autres champs disciplinaires et consulter aussi par exemple les travaux des sémioticiens ou des historiens de l'écriture[32].

Mais quelle que soit l'essence des diagrammes, notre étude a révélé que les systèmes inférentiels diagrammatiques et les systèmes logico-linguistiques avaient des propriétés distinctes, souvent duales et antinomiques. Et c'est finalement une excellente nouvelle, car ces propriétés peuvent être intelligemment combinées dans des systèmes de représentation hétérogènes comportant plusieurs modules. Ce nouveau cadre de recherche ouvre de nombreuses perspectives d'applications pratiques, et nous avons consacré la dernière partie aux perspectives d'application des SRH dans deux domaines des sciences cognitives : celui de la modélisation du raisonnement et celui de la sémantique du langage naturel[33]. Mais il y a en réalité de nombreuses applications en I.A., que ce soit en robotique, en vision par ordinateur et, plus généralement, dans la programmation de systèmes complexes. Dans des systèmes intelligents, des représentations diagrammatiques pourraient même être paradoxalement utilisées pour représenter des relations du second ordre, diminuer la complexité des calculs ou les orienter.

---

[32] On peut citer ici les travaux de Jack Goody ou Anne-marie Christin notamment (cf. Goody 1979, Christin 2001).

[33] Je remercie Dick Carter et Patricia Zablit qui m'ont conduite en 1993 à explorer ce domaine en me communiquant leur enthousiasme pour les représentations analogiques.

# *Références bibliographiques*

Barwise J. et Etchemendy J., 1990 a, « Information, Infons, and Inference», dans *Situation Theory and its applications*, vol. 1, Lecture Notes no. 22, CSLI, Stanford.
Barwise J. et Etchemendy J., 1990 b, « Visual Information and Valid Reasoning», dans *Visualization in Mathematics*, Zimmerman, W., ed., Mathematical Association of America, Washington DC, et dans *Phylosophy and the Computer*, Westview Press, 1992.
Barwise J. et Etchemendy J., 1994, *Hyperproof.* Stanford : CSLI Publications.
Barwise J. et Etchemendy J., « Heterogenous Logic», 1995, dans Glasgow et alii 1995 (présenté initialement en 1993 à IJCAI, dans le Workshop 'Principle of hybrid representation and reasoning', Chamberry, France).
Brachman R. et Levesque H., 1985, « A fundamental tradeoff in knowledge representation and reasoning » dans *Readings in Knowledge Representation*, Morgan Kaufman.
Bruner J., Goodnow J. & Austin A., 1956, A study of thinking, Wiley, New York.
Carter R., Recanati C. et Zablit P., fev. 1993, "Modélisation cognitive de la représentation du temps et de l'espace dans la compréhension et la production du langage", Proposition Technique, Programme de Recherches CogniSeine.
Couturat L., 1903, Opuscules et fragments inédits de Leibniz, F. Alcan, Paris.
Christin A-M., 2001, *Histoire de l'écriture, de l'idéogramme au multimédia*, textes réunis et présentés par Anne-Marie Christin, Edition Flammarion.
Dretske F., 1981, *Knowledge and the flow of information*, Blackwell, Oxford.
Euler L., 1768-1772, Lettres à une princesse d'Allemagne sur quelques sujets de physique et de philosophie. Deutsche Übersetzung: Leipzig 1773-1780 et dans Speiser A., Trost E. et Blanc C. (Eds.), 1960, *Œuvres Complètes d'Euler*, Orell Füssl., Zurich.
Fauconnier G., 1997, *Mappings in thought and language*, Cambridge University Press.
Gentner D., Holyoak K. et alii (eds), 2001, *The analogical mind*, MIT Press, Cambridge, Massachusetts and London, England.
Glasgow J., Nari Narayanan N., Chandrasekaran B., eds, 1995, *Diagrammatic reasoning : cognitive and computational perspective*, AAAI Press/MIT Press, Cambridge, Massachusetts and London, England.
Goody J., 1979, *La raison graphique: la domestication de la pensée sauvage*, collection Le sens commun, Editions de Minuit, Paris.
Gosselin L., 1996, *Sémantique de la temporalité en français. Un modèle calculatoire et cognitif du temps et de l'aspect*, Duculot, Louvain-la-Neuve.
Guillaume G., 1929, *Temps et Verbe, Théorie des aspects, des modes et des temps*, Collection Linguistique publiée par la Société de Linguistique de Paris, 27.
Hammer E., 1994, « Reasoning with sentences and diagrams » dans Notre Dame Journal of Formal Logic, vol. 35, nb 1, Winter 1994.
Hornstein N., 1993, *As Time Goes By - Tense and Universal Grammar*, A Bradford book, MIT Press, Cambridge Mass.
Johnson-Laird P.N., 1983, *Mental Models: towards a cognitive science of language, inference, and consciousness*, Cambridge University Press, Cambridge.
Kayser D., 1992, « Intelligence Artificielle et Modélisation Cognitive : objectifs et évaluation », Intellectica n°13-14, pp. 223-240.
Langacker R. W., 1987, *Foundations of Cognitive Grammar,* vol 1, Stanford University Press, California.
Langacker R. W., 1991, *Concept, Image, and Symbol - The Cognitive Basis of Grammar* - Coll. Cognitive Linguistics Research, Mouton de Gruyter.
Larkin J., et Simon H., 1987, « Why a Diagram Is (Sometimes) Worth Ten Thousand Words»,

*Cognitive Science*, vol 11.
Macken E., Perry J., et Hass C., 1993, « Richly Grounding Symbols in ASL», CSLI Report no. 93-180, Sep. 1993 (paru également dans la revue *Sign Language Studies* en dec. 93).
Marr D., *Vision*, 1982, Freeman W.H., San Francisco, 1982.
Nicod J., 1962, *La géométrie dans le monde sensible*, Bibliothèque de Philosophie contemporaine fondée par Félix Alcan, PUF.
Peirce C. S., 1933, *Collected Papers*, vol. 4, Hartshorne C. & Weiss P., eds, Cambdridge, MA : Harvard University Press.
Perry J. et Macken E., 1996, « Interfacing Situations », dans *Logic, Language and Computation*, Seligmann J. and Westerstahl D., eds, Stanford University.
Recanati C., 2004, « Diagrammes pour résoudre le problème d'Einstein et celui d'un joueur de MasterMind », rapport LIPN, Dec. 2004, Université Paris13.
Reichenbach, H., 1947, *Elements of symbolic Logic*, Macmillan, New York.
Shin, S-J., 1991, « A Situation-Theoretic Account of Valid Reasoning with Venn Diagrams» dans *Situation Theory and its applications*, vol. 2, Lecture Notes no. 26, CSLI, Stanford.
Shin, S-J., 1994, *The logical status of Diagrams*, Cambridge University Press.
Sowa, J., 1984, *Conceptual Structure: Information Processing in Mind and Machine*, Addison-Wesley, Reading, MA.
Sloman, A., 1995, « Musings on the role of logical and nonlogical representations in intelligence», in *Diagrammatic Reasoning : cognitive and computational perspective*, J. Glasgow, N. Nari Narayanan, B. Chandrasekaran, eds, AAAI Press/MIT Press, Cambridge, Massachusetts and London, England.
Smith, C., 1991, *The parameter of Aspect*, Studies in Linguistics and Philosophy 43, Kluwer Academic Publishers.
Sperber D. et Wilson D., 1986, *Relevance: Communication and Cognition*, Basil Blackwell, Oxford.
Stenning K. et Oberlander I., 1991, « Reasoning with Words, Pictures, and calculi: Computation Versus Justification» dans *Situation Theory and its applications*, vol. 2, Lecture Notes no. 26, CSLI, Stanford.
Stenning K. et Oberlander I., 1995, « A Cognitive Theory of Graphical and Linguistic Reasoning: Logic and Implementation» in Cognitive Science, vol. 19 (1).
Tennant N., 1986, « The withering away of Formal Semantics » in *Mind and Language* 1 : 302-318.
Tversky B., 2003, « Spatial Schemas in Depictions », in Gattis M. (ed), *Spatial schemas and abstract thought*, Cambridge, MIT Press.
Vandeloise C., 1986, *L'espace en français : sémantique des prépositions spatiales*, Travaux en linguistique, Editions du Seuil, Paris.
Venn J., *Symbolic Logic*, 1894-1971, Chelsea Publishing Company, New York.
Vieu L., 1991, *Sémantique des relations spatiales et inférences spatio-temporelles : une contribution à l'étude des structures formelles de l'espace en langage naturel*, Thèse de doctorat de l'Université Paul Sabatier, Toulouse.
Zablit P., 1991, *Construction de l'interprétation temporelle en langue naturelle*. Thèse de Doctorat, Université Paris 11, Orsay.